%% file: main.tex
\documentclass{article}
\usepackage{ijcai24}

\usepackage{times}
\usepackage{soul}
\usepackage{url}
\usepackage[hidelinks]{hyperref}
\usepackage[utf8]{inputenc}
\usepackage[small]{caption}
\usepackage{graphicx}
\usepackage{amsmath}
\usepackage{amsthm}
\usepackage{booktabs}
\usepackage{algorithm}
\usepackage[noend]{algorithmic}
\usepackage[switch]{lineno}

\newtheorem{theorem}{Theorem}

\input{macros}

\title{Proximal Curriculum with Task Correlations \\ for Deep Reinforcement Learning}

\author{
Georgios Tzannetos$^1$
\and
Parameswaran Kamalaruban$^2$\And
Adish Singla$^1$
\affiliations
$^1$Max Planck Institute for Software Systems\\
$^2$The Alan Turing Institute
\emails
\{gtzannet, adishs\}@mpi-sws.org,
kparameswaran@turing.ac.uk
}

\begin{document}

\maketitle

\input{0_abstract}

\input{1_introduction}

\input{2_problem_setup}

\input{3_method}

\input{4_experiments}

\input{5_conclusion}

%\appendix
%\section*{Ethical Statement}

%%%%%%%%%%%%%%%%%%%%%%%%%%%%%%%%%%%%%%%%%%%%%%%%%%%%%%%%%
\section*{Acknowledgments}
Parameswaran Kamalaruban acknowledges support from The Alan Turing Institute.
Funded/Co-funded by the European Union (ERC, TOPS, 101039090). Views and opinions expressed are however those of the author(s) only and do not necessarily reflect those of the European Union or the European Research Council. Neither the European Union nor the granting authority can be held responsible for them.
%%%%%%%%%%%%%%%%%%%%%%%%%%%%%%%%%%%%%%%%%%%%%%%%%%%%%%%%%

\bibliography{main}
\bibliographystyle{named}

\clearpage    
\onecolumn
\input{9_appendix}

\end{document}

%% file: macros.tex
\usepackage{subcaption}
\usepackage{wrapfig} % Forbidden in aaai
\usepackage{etoolbox}
\usepackage{verbatim}
\usepackage[inline]{enumitem}

\usepackage{makecell}
\usepackage{amssymb}
\usepackage{ragged2e}
\usepackage{nameref}

\newtheorem{proposition}{Proposition}

\DeclareMathOperator*{\argmax}{arg\,max}

\newcommand{\w}{\ensuremath{\theta_t}}
\newcommand{\wnext}{\ensuremath{\theta_{t+1}}}

\newcommand\norm[1]{\left\lVert#1\right\rVert}

\newcommand{\bcc}[1]{\left\{{#1}\right\}}
\newcommand{\brr}[1]{\left({#1}\right)}
\newcommand{\bss}[1]{\left[{#1}\right]}
\newcommand{\ipp}[2]{\left\langle{#1},{#2}\right\rangle}

\newcommand{\Expect}[1]{\mathbb{E}\bss{{#1}}}
\newcommand{\Expectover}[2]{\mathbb{E}_{#1}\!\left[#2\right]}

\newcommand{\envSGR}{\textsc{SGR}{}}
\newcommand{\envPointMassS}{\textsc{PM-s}{}}
\newcommand{\envBipedalWalker}{\textsc{BW}{}}

\newcommand{\AlgoOurs}{\textsc{ProCuRL-Target}{}}
\newcommand{\AlgoOursUn}{\textsc{ProCuRL-Unif}{}}

\usepackage{thmtools}
\usepackage{thm-restate}

%% file: 0_abstract.tex
% !TEX root = main.tex
%%%%%%%%%%%%%%%%%%%%%%%%%%%%%%%%%%%%%
%%%%%%%%%%%%%%%%%%%%%%%%%%%%%%%%%%%%%
\begin{abstract}
Curriculum design for reinforcement learning (RL) can speed up an agent's learning process and help it learn to perform well on complex tasks. However, existing techniques typically require domain-specific hyperparameter tuning, involve expensive optimization procedures for task selection, or are suitable only for specific learning objectives. In this work, we consider curriculum design in contextual multi-task settings where the agent's final performance is measured w.r.t. a \emph{target distribution over complex tasks}. We base our curriculum design on the Zone of Proximal Development concept, which has proven to be effective in accelerating the learning process of RL agents for \emph{uniform distribution over all tasks}. We propose a novel curriculum, \AlgoOurs{}, that effectively balances the need for selecting tasks that are not too difficult for the agent while progressing the agent's learning toward the target distribution via leveraging task correlations. We theoretically justify the task selection strategy of \AlgoOurs{} by analyzing a simple learning setting with \textsc{Reinforce} learner model. Our experimental results across various domains with challenging target task distributions affirm the effectiveness of our curriculum strategy over state-of-the-art baselines in accelerating the training process of deep RL agents.
\end{abstract}

%% file: 1_introduction.tex
% !TEX root =  main.tex
%%%%%%%%%%%%%%%%%%%%%%%%%%%%%%%%%%%%%
%%%%%%%%%%%%%%%%%%%%%%%%%%%%%%%%%%%%%
\section{Introduction} \label{sec:intro}

Deep reinforcement learning (RL) has shown remarkable success in various fields such as games, continuous control, and robotics, as evidenced by recent advances in the field~\cite{mnih2015human,lillicrap2015continuous,silver2017mastering,levine2016end}. However, despite these successes, the broader application of RL in real-world domains is often very limited. Specifically, training RL agents in complex environments, such as contextual multi-task settings and goal-based tasks with sparse rewards, still presents significant challenges~\cite{kirk2021survey,andrychowicz2017hindsight,florensa2017reverse,riedmiller2018learning}.

Curriculum learning has been extensively studied in the context of supervised learning~\cite{weinshall2018curriculum,zhou2018minimax,elman1993learning,bengio2009curriculum}. Recent research has explored the benefits of using curriculum learning in sequential decision making settings, such as reinforcement learning and imitation learning~\cite{florensa2017reverse,riedmiller2018learning,wohlke2020performance,florensa2018automatic,racaniere2019automated,klink2020selfdeep,eimer2021self,imt_ijcai2019,yengera2021curriculum}. The objective of curriculum design in RL is to speed up an agent's learning process and enable it to perform well on complex tasks by exposing it to a personalized sequence of tasks~\cite{narvekar2020curriculum,portelas2021automatic}. To achieve this objective, several works have proposed different curriculum strategies based on different design principles, such as the Zone of Proximal Development (ZPD)~\cite{vygotsky1978mind,chaiklin2003analysis}, Self-Paced Learning (SPL)~\cite{kumar2010self,jiang2015self}, and Unsupervised Environment Design (UED)~\cite{dennis2020emergent}. However, existing techniques typically require domain-specific hyperparameter tuning, involve expensive optimization procedures for task selection, or are suitable only for specific learning objectives, such as uniform performance objectives.

In this work, we investigate curriculum design in contextual multi-task settings with varying degrees of task similarity, where the agent's final performance is measured w.r.t. a \emph{target distribution over complex tasks}. We base our curriculum design on the Zone of Proximal Development concept, which has proven to be effective in accelerating the learning process of RL agents for \emph{uniform distribution over all tasks}~\cite{florensa2017reverse,wohlke2020performance,florensa2018automatic,tzannetos2023proximal}. We propose a novel curriculum strategy, \AlgoOurs{}, that effectively balances the need for selecting tasks that are neither too hard nor too easy for the agent (according to the ZPD concept) while still progressing its learning toward the target distribution via leveraging task correlations. We have mathematically derived our curriculum strategy by analyzing a specific learning setting. The strengths of our curriculum strategy include its broad applicability to many domains with minimal hyperparameter tuning, computational and sample efficiency, easy integration with deep RL algorithms, and applicability to any target distribution over tasks, not just uniform distribution. Our main results and contributions are as follows:
\begin{enumerate}[label={\Roman*.},leftmargin=*]
\item We propose a curriculum strategy, \AlgoOurs{}, that effectively trades off the suitable task difficulty level for the agent and the progression towards the target tasks (Section~\ref{sec:prox-corl-curr}). 
\item We mathematically derive \AlgoOurs{} for the single target task setting with a discrete pool of tasks by analyzing the effect of picking a task on the agent's learning progress (Section~\ref{sec:curr-discrete-single}). 
\item We propose an extension of \AlgoOurs{} that can be applied to a wide range of task spaces and target distributions. This extension can be seamlessly integrated with deep RL frameworks, making it easy to use and apply in various scenarios (Section~\ref{sec:prac-algs}).
\item We empirically demonstrate that the curricula generated with \AlgoOurs{} significantly improve the training process of deep RL agents in various environments, matching or outperforming existing state-of-the-art baselines (Section~\ref{sec:experiments}).\footnote{Github repository: \url{https://github.com/machine-teaching-group/ijcai2024-proximal-curriculum-target-rl}\label{footnote:gitcode}}
\end{enumerate}

\subsection{Related Work}
\label{subsec:detailed-comparison}

\looseness-1\textbf{Curriculum strategies based on SPL concept.} In the realm of supervised learning, curriculum strategies leveraging the SPL concept attempt to strike a balance between exposing the learner to all available training examples and selecting examples in which it currently
performs well~\cite{kumar2010self,jiang2015self}. In the context of RL, the SPL concept has been adapted by researchers in \textsc{SPDL}~\cite{klink2020selfdeep,klink2021probabilistic}, \textsc{SPaCE}~\cite{eimer2021self}, and \textsc{CURROT}~\cite{klink2022curriculum} by controlling the intermediate task distribution with respect to the learner's current training progress. While both \textsc{SPDL} and \textsc{CURROT} involve a setting where the learner's performance is measured w.r.t. a target distribution over the task space (similar to our objective), \textsc{SPaCE} operates in a setting where the learner's performance is measured w.r.t. a uniform distribution over the task space. 
The task selection mechanism varies across these methods. \textsc{SPDL} and \textsc{CURROT} operate by solving an optimization problem at each step to select the most relevant task~\cite{klink2021probabilistic,klink2022curriculum}. On the other hand, \textsc{SPaCE} relies on ranking tasks based on the magnitude of differences in current/previous critic values to choose the task for the next step~\cite{eimer2021self}. Furthermore, the work of \textsc{CURROT}~\cite{klink2022curriculum} showcases issues about using KL divergence to measure the similarity between task distributions as used in \textsc{SPDL} -- instead, they introduce an alternative approach by posing the curriculum design as a constrained optimal transport problem between task distributions.

\looseness-1\textbf{Curriculum strategies based on UED concept.} The UED problem setting involves automatically designing a distribution of environments that adapts to the learning agent~\cite{dennis2020emergent}. UED represents a self-supervised RL paradigm in which an environment generator evolves alongside a student policy to develop an adaptive curriculum learning approach. This approach can be utilized to create increasingly complex environments for training a policy, leading to the emergence of Unsupervised Curriculum Design. \textsc{PAIRED}~\cite{dennis2020emergent} is an adversarial training technique that solves the problem of the adversary generating unsolvable environments by introducing an antagonist who works with the environment-generating adversary to design environments in which the protagonist receives a low reward. Furthermore, the connections between UED and another related method called \textsc{PLR}~\cite{jiang2021prioritized} have been explored in~\cite{jiang2021replay,parker2022evolving}, resulting in demonstrated improvements over \textsc{PAIRED}. \textsc{PLR}, originally designed for procedural content generation based environments, samples tasks/levels by prioritizing those with higher estimated learning potential when revisited in the future. TD errors are used to estimate a task’s future learning potential. Unlike~\cite{jiang2021replay,parker2022evolving}, \textsc{PLR} does not assume control over the environment generation process, requiring only a black box generation process that returns a task given an identifier.

\looseness-1\textbf{Curriculum strategies based on ZPD concept.} Effective teaching provides tasks of moderate difficulty (neither too hard nor too easy) for the learner, as formalized by the ZPD concept~\cite{vygotsky1978mind,chaiklin2003analysis,oudeyer2007intrinsic,baranes2013active,zou2019towards}. In the context of RL, several curriculum strategies are based on the ZPD concept, such as selecting the next task randomly from a set of tasks with success rates within a specific range~\cite{florensa2017reverse,florensa2018automatic}. However, the threshold values for success rates require tuning based on the learner's progress and domain. A unified framework for performance-based starting state curricula in RL is proposed by \cite{wohlke2020performance}, while \cite{tzannetos2023proximal} propose a broadly applicable ZPD-based curriculum strategy with minimal hyperparameter tuning and theoretical justifications. Nonetheless, these techniques are generally suitable only for settings where the learner's performance is evaluated using a uniform distribution over all tasks.

\looseness-1\textbf{Curriculum strategies based on domain knowledge.} In supervised learning, early works involve ordering examples by increasing difficulty~\cite{elman1993learning,bengio2009curriculum,schmidhuber2013powerplay}, which has been adapted in hand-crafted RL curriculum approaches~\cite{wu2016training}. Recent works on imitation learning have also utilized iterative machine teaching framework to design greedy curriculum strategies~\cite{imt_ijcai2019,yengera2021curriculum,liu2017iterative,DBLP:journals/corr/ZhuSingla18}. However, these approaches require domain-specific expert knowledge.
%for designing difficulty measures.

\textbf{Other automatic curriculum strategies.} Various automatic curriculum generation approaches exist, including: (i) formulating the curriculum design problem as a meta-level Markov Decision Process~\cite{narvekar2019learning}; (ii) learning to generate training tasks similar to a teacher~\cite{dendorfer2020goal,matiisen2019teacher,turchetta2020safe}; (iii) using self-play for curriculum generation~\cite{sukhbaatar2018intrinsic}; (iv) leveraging disagreement between different agents trained on the same tasks~\cite{zhang2020automatic}; and (v) selecting starting states based on a single demonstration~\cite{salimans2018learning}. Interested readers can refer to recent surveys on RL curriculum design~\cite{narvekar2020curriculum,portelas2021automatic}.
%~\cite{salimans2018learning,resnick2018backplay}. 

%% file: 2_problem_setup.tex
% !TEX root = main.tex
%%%%%%%%%%%%%%%%%%%%%%%%%%%%%%%%%%%%%%%%%%%%%%%%%%%%%%%%
%%%%%%%%%%%%%%%%%%%%%%%%%%%%%%%%%%%%%%%%%%%%%%%%%%%%%%%%

\begin{algorithm*}[t]
    \caption{RL Agent Training as Interaction between Teacher-Student Components}
    \begin{algorithmic}[1]
        \STATE \textbf{Input:} RL agent's initial policy $\pi_{1}$
        \FOR{$t = 1,2,\dots$}
            \STATE Teacher component picks a task $c_t \in \mathcal{C}$.  \label{alg1:line_curriculum}
            \STATE Student component attempts the task via a trajectory rollout $\xi_t$ using the policy $\pi_t$ in $\mathcal{M}_{c_t}$.
            \STATE Student component updates the policy to $\pi_{t+1}$ using the rollout $\xi_t$. \label{alg1:line_update}
        \ENDFOR{}
        \STATE \textbf{Output:} RL agent's final policy $\pi_{\textnormal{end}} \gets \pi_{t+1}$. \label{alg1:line_output}
    \end{algorithmic}
    \label{alg:interaction}
\end{algorithm*}

\section{Formal Setup}
\label{sec:formal-setup}

We formalize our problem setting based on prior work on teacher-student curriculum learning~\cite{matiisen2019teacher}. 

\textbf{Multi-task RL.} We consider a multi-task RL setting with a task/context space $\mathcal{C}$, in which each task $c \in \mathcal{C}$ is associated with a learning environment modeled as a contextual Markov Decision Process (MDP), denoted by $\mathcal{M}_c := \brr{\mathcal{S}, \mathcal{A}, \gamma, \mathcal{T}_c, R_c, P^0_c}$~\cite{hallak2015contextual,modi2018markov}. The state space $\mathcal{S}$ and action space $\mathcal{A}$ are shared by all tasks in $\mathcal{C}$, as well as the discount factor $\gamma$. Each contextual MDP includes a contextual transition dynamics $\mathcal{T}_c: \mathcal{S} \times \mathcal{S} \times \mathcal{A} \rightarrow \bss{0,1}$, a contextual reward function $R_c: \mathcal{S} \times \mathcal{A} \rightarrow \bss{-R_{\mathrm{max}}, R_{\mathrm{max}}}$, where $R_{\mathrm{max}} > 0$, and a contextual initial state distribution $P^0_c: \mathcal{S} \rightarrow \bss{0,1}$. We denote the space of environments by $\mathcal{M} = \bcc{\mathcal{M}_c: c \in \mathcal{C}}$. Moreover, we have a target distribution $\mu$ over $\mathcal{C}$ that is used for performance evaluation, as further discussed below.

\textbf{RL agent and training process.} We consider an RL agent acting in any environment $\mathcal{M}_c \in \mathcal{M}$ via a contextual policy $\pi: \mathcal{S} \times \mathcal{C} \times \mathcal{A} \rightarrow \bss{0,1}$ that is a contextual mapping from a state to a probability distribution over actions. Given a task $c \in \mathcal{C}$, the agent attempts the task via a trajectory rollout obtained by executing its policy $\pi$ in the MDP $\mathcal{M}_c$. The trajectory rollout is denoted as $\xi = \bcc{(s^{(\tau)},a^{(\tau)})}_{\tau = 0,1,\dots}$ with $s^{(0)} \sim P^0_c$. The agent's performance on task $c$ is measured by the value function $V^\pi (c) := \mathbb{E}\bss{\sum_{\tau=0}^\infty \gamma^\tau \cdot R_c(s^{(\tau)},a^{(\tau)}) \big| \pi, \mathcal{M}_c}$. The agent training corresponds to finding a policy that performs well w.r.t. the target distribution $\mu$, i.e., $\max_\pi V^\pi_\mu$ where $V^\pi_\mu := \mathbb{E}_{c \sim \mu} \bss{V^\pi (c)}$. The training process of the agent involves an interaction between two components: a student component that is responsible for policy updates and a teacher component that is responsible for task selection. The interaction happens in discrete steps indexed by $t=1, 2, \ldots$, and is formally described in Algorithm~\ref{alg:interaction}. Let $\pi_{\textnormal{end}}$ denote the agent's final policy at the end of teacher-student interaction. The \emph{training objective} is to ensure that the performance of the policy $\pi_{\textnormal{end}}$ is $\epsilon$-near-optimal, i.e., $(\max_\pi V^\pi_\mu - V^{\pi_{\textnormal{end}}}_\mu) \leq \epsilon$.

\looseness-1\textbf{Student component.} We consider parametric policies of the form $\pi_\theta : \mathcal{S} \times \mathcal{C} \times \mathcal{A} \rightarrow \bss{0,1}$, where $\theta \in \Theta \subseteq \mathbb{R}^d$. The agent's policy at step $t$ is given by $\pi_{t} := \pi_{\theta_{t}}$. The student component updates the policy parameter based on the following quantities: the current parameter $\theta_t$, the task $c_t$ picked by the teacher component, and the rollout $\xi_t = \bcc{(s^{(\tau)}_t,a^{(\tau)}_t)}_{\tau}$. For example, the policy parameter of the \textsc{Reinforce} agent~\cite{sutton1999policy} is updated as follows: $\wnext \gets \w + \eta_t \cdot \sum_{\tau = 0}^{\infty} G_t^{(\tau)} \cdot \bss{\nabla_\theta \log \pi_\theta (a_t^{(\tau)}|s_t^{(\tau)}, c_t)}_{\theta = \theta_t}$, where $\eta_t$ is the learning rate, and $G_t^{(\tau)} = \sum_{\tau' = \tau}^\infty \gamma^{\tau' - \tau} \cdot R_{c_t} (s_t^{(\tau')}, a_t^{(\tau')})$. 

\textbf{Teacher component.} At step $t$, the teacher component selects a task $c_t$ for the student component to attempt via a trajectory rollout, as shown in line~\ref{alg1:line_curriculum} in Algorithm~\ref{alg:interaction}. The sequence of tasks, also known as the curriculum, that is chosen by the teacher component has a significant impact on the performance improvement of the policy $\pi_t$. The primary objective of this work is to develop a teacher component to achieve the training objective in a computationally efficient and sample-efficient manner.

%% file: 3_method.tex
% !TEX root = main.tex
%%%%%%%%%%%%%%%%%%%%%%%%%%%%%%%%%%%%%%%%%%%%%%%%%%%%%%%%
%%%%%%%%%%%%%%%%%%%%%%%%%%%%%%%%%%%%%%%%%%%%%%%%%%%%%%%%
\section{Our Curriculum Strategy}
\label{sec:prox-corl-curr}
% \AlgoOurs{}

In Section~\ref{sec:curr-discrete-single}, we mathematically derive a curriculum strategy for the single target task setting with a discrete pool of tasks. Then, in Section~\ref{sec:prac-algs}, we present our final curriculum strategy that is applicable in general learning settings. The proofs are provided in appendices of the paper.

\subsection{Curriculum Strategy for Single Target Settings}
\label{sec:curr-discrete-single}

In this section, we present our curriculum strategy for a setting where the task space $\mathcal{C}$ is a discrete set and the target distribution $\mu$ is a delta distribution concentrated on a single target task $c_\textnormal{targ}$. To design our curriculum strategy, we investigate the effect of selecting a task $c_t$ at time step $t$ on the agent's performance $V^{\pi_{\theta_t}}_\mu$ and its convergence towards the target performance $V^{*}_\mu := \max_\pi V^{\pi}_\mu$. Therefore, we define the training objective improvement at step $t$ and analyze this metric across both general and specific learning scenarios.

\textbf{Expected improvement in the training objective.} At step $t$, given the current policy parameter $\theta_t$, the task $c_t$ picked by the teacher component, and the student component's rollout $\xi_t$, we define the improvement in the training objective as: 
\begin{equation*}
\Delta_t (\theta_{t+1} \big| \theta_t, c_t, \xi_t) ~:=~ (V^{*}_\mu - V^{\pi_{\theta_t}}_\mu) - (V^{*}_\mu - V^{\pi_{\theta_{t+1}}}_\mu) .
\end{equation*}
Additionally, we define the expected improvement in the training objective at step $t$ due to picking the task $c_t$ as follows~\cite{weinshall2018curriculum,imt_ijcai2019,yengera2021curriculum,graves2017automated}:
\begin{equation*}
I_t (c_t) ~:=~ \Expectover{\xi_t \mid c_t}{\Delta_t (\theta_{t+1} | \theta_t, c_t, \xi_t)} .
\end{equation*}
Based on the above measure, a natural greedy curriculum strategy for selecting the next task $c_t$ is given by: 
\begin{equation}
c_t ~\gets~ \argmax_{c \in \mathcal{C}} I_t(c) .
\label{eq:natural-greedy-curriculum}
\end{equation}
We aim to approximate such a curriculum strategy without computing the updated policy $\pi_{\theta_{t+1}}$. To this end, we initially analyze the function $I_t(\cdot)$ for \textsc{Reinforce} learner model within a general learning setting. Subsequently, we refine and simplify the findings by delving into a specific learning scenario. This analysis enables us to develop an intuitive curriculum strategy by effectively combining the following fundamental factors: (i) the learning potential inherent in the source and target tasks, and (ii) the transfer potential between the source and target tasks, i.e., their similarity.

\textbf{Gradient alignment approximation of $I_t(\cdot)$.} Here, we analyze the function $I_t(\cdot)$ within the context of the \textsc{Reinforce} learner model operating under a general learning setting. We show that the natural greedy curriculum strategy can be approximated by a simple gradient alignment maximization strategy. Initially, through the application of the first-order Taylor approximation of $V^{\pi_{\theta_{t+1}}} (c_\textnormal{targ})$ at $\theta_t$, we approximate the improvement in the training objective as follows:
\begin{align*} 
\Delta_t (\theta_{t+1} \big| \theta_t, c_t, \xi_t) ~=~& V^{\pi_{\theta_{t+1}}} (c_\textnormal{targ}) - V^{\pi_{\theta_t}} (c_\textnormal{targ}) \\ 
~\approx~& \ipp{\theta_{t+1} - \theta_t}{g_t(c_\textnormal{targ})} ,
\end{align*}
where $g_t(c) := \bss{\nabla_\theta V^{\pi_{\theta}} (c)}_{\theta = \theta_t}$. Subsequently, by utilizing the parameter update form of the \textsc{Reinforce} agent (i.e., $\Expectover{\xi_t \mid c_t}{\theta_{t+1}} \gets \theta_t + \eta_t \cdot g_t(c_t)$), we approximate the expected improvement in the training objective as follows:
\begin{align*}
I_t(c_t) ~\approx~& \ipp{\Expectover{\xi_t \mid c_t}{\theta_{t+1} - \theta_t}}{g_t(c_\textnormal{targ})} \\
~=~& \eta_t \cdot \ipp{g_t(c_t)}{g_t(c_\textnormal{targ})} .
\end{align*}
Consequently, the natural greedy curriculum strategy in Eq.~\eqref{eq:natural-greedy-curriculum} can be effectively approximated by the following gradient-alignment-based curriculum strategy:
\begin{equation}
c_t ~\gets~ \argmax_c \ipp{g_t(c)}{g_t(c_\textnormal{targ})} . \label{eq:curr-gradient-form}
\end{equation}
In the following theorem, we demonstrate the effectiveness of employing the above curriculum strategy in accelerating the convergence of the \textsc{Reinforce} agent.
\begin{theorem}
\label{thm:curriculum-general}
Consider Algorithm~\ref{alg:interaction} with the \textsc{Reinforce} learner model and the curriculum strategy defined in Eq.~\eqref{eq:curr-gradient-form}. Then, after $t = \mathcal{O}\brr{\log \frac{1}{\epsilon}}$ steps, we have: 
\[
\Expect{(V^{*} (c_\textnormal{targ}) - V^{\pi_{\theta_t}} (c_\textnormal{targ})) \mid \theta_0} \leq \epsilon , 
\]
where $V^*(c) := \max_\pi V^\pi(c)$.
\end{theorem}
Subsequently, building on the curriculum strategy outlined in Eq.~\eqref{eq:curr-gradient-form}, we devise an intuitive curriculum strategy through an analysis of the curriculum objective $\ipp{g_t(c)}{g_t(c_\textnormal{targ})}$ within the context of a contextual bandit setting.

\textbf{Further simplification of gradient alignment.} We consider the \textsc{Reinforce} learner model with the following policy parameterization: given a feature mapping $\phi: \mathcal{S} \times \mathcal{C} \times \mathcal{A} \to \mathbb{R}^d$, for any $\theta \in \mathbb{R}^d$, we parameterize the policy as $\pi_\theta (a | s, c) = \frac{\exp (\ipp{\theta}{\phi(s,c,a)})}{\sum_{a'} \exp (\ipp{\theta}{\phi(s,c,a')})}, \forall s \in \mathcal{S}, c \in \mathcal{C}, a \in \mathcal{A}$. In the following, we consider a specific problem instance of contextual MDP setting. Let $\mathcal{M}_c$ be a contextual MDP with a singleton state space $\mathcal{S} = \bcc{s}$, and an action space $\mathcal{A} = \bcc{a_1, a_2}$. Any action $a \in \mathcal{A}$ taken from the initial state $s \in \mathcal{S}$ always leads to a terminal state. Let $r: \mathcal{C} \to [0,1]$ be a mapping from task/context space $\mathcal{C}$ to the interval $[0,1]$. For any context $c \in \mathcal{C}$, we denote the optimal and non-optimal actions for that context as $a_c^\textnormal{opt}$ and $a_c^\textnormal{non}$, respectively. The contextual reward function is defined as follows: $R_c(s,a_c^\textnormal{opt}) = 1$, and $R_c(s,a_c^\textnormal{non}) = 0$, for all $c \in \mathcal{C}$. Further, we define $\psi: \mathcal{C} \rightarrow \mathbb{R}^d$ as $\psi(c) := (\phi(s, c, a_c^\textnormal{opt}) - \phi(s, c, a_c^\textnormal{non}))$. Subsequently, for the \textsc{Reinforce} agent operating under the above setting, the following proposition quantifies the objective term of the curriculum strategy as per Eq.~\eqref{eq:curr-gradient-form} at step $t$:
\begin{proposition}
\label{prop:prox-analysis-reinforce-learner}
For the \textsc{Reinforce} agent with softmax policy parameterization under the contextual bandit setting described above, we have:
\begin{align*}
\ipp{g_t(c)}{g_t(c_\textnormal{targ})} ~=~& \eta_t \cdot Z_t(c) \cdot Z_t(c_\textnormal{targ}) \cdot \ipp{\psi(c)}{\psi(c_\textnormal{targ})} ,
\end{align*}
where $Z_t(c) := \frac{V^{\pi_{\theta_t}} (c)}{V^* (c)} \cdot \big(V^* (c) - V^{\pi_{\theta_t}} (c)\big)$ denotes the agent's learning potential on task $c$ at step $t$.  
\end{proposition} 

\textbf{Our curriculum strategy.} Inspired by the above analysis, we propose the following curriculum strategy: 
\begin{align}
& c_t~\gets~\argmax_{c \in \mathcal{C}} \underbrace{Z_t(c)}_{\textcircled{A}} \cdot \underbrace{Z_t(c_\textnormal{targ})}_{\textcircled{B}} \cdot \underbrace{\ipp{\psi(c)}{\psi(c_\textnormal{targ})}}_{\textcircled{C}}, 
\label{eq:cta-scheme-rl}
\end{align}
where $\psi: \mathcal{C} \to \mathbb{R}^d$ is a context representation mapping. At step $t$, the teacher component picks a task $c_t$ according to Eq.~\eqref{eq:cta-scheme-rl}. The curriculum strategy involves the following quantities: $\textcircled{A}$ the agent's learning potential on task $c$, $\textcircled{B}$ the agent's learning potential on task $c_\textnormal{targ}$, and $\textcircled{C}$ the correlation between the tasks $c$ and $c_\textnormal{targ}$. Term $\textcircled{A}$ enforces the selection of tasks that are neither too hard nor too easy for the current policy, aligning with the ZPD principle. The combined effect of terms $\textcircled{B}$ and $\textcircled{C}$ emphasizes the choice of tasks highly correlated with the target task (which has high learning potential). The curriculum strategy effectively balances these two objectives.   

%%%%%%%%%%
\input{figs/environments/fig_environments}
%%%%%%%%%%

\subsection{Curriculum Strategy for General Settings}
\label{sec:prac-algs}

In this section, we extend the curriculum strategy in Eq.~\eqref{eq:cta-scheme-rl} to practical settings of interest, i.e., a general task space $\mathcal{C}$, a general target distribution $\mu$, and $V^* (c)$ values being unknown. We begin by constructing two large discrete sets, $\widehat{\mathcal{C}}_\textnormal{unif}$ and $\widehat{\mathcal{C}}_\textnormal{targ}$, which are subsets of the original task space $\mathcal{C}$. $\widehat{\mathcal{C}}_\textnormal{unif}$ is obtained by sampling contexts from $\mathcal{C}$ according to uniform distribution, while $\widehat{\mathcal{C}}_\textnormal{targ}$ is obtained by sampling contexts from $\mathcal{C}$ according to the target distribution $\mu$. For the general setting, we consider the following curriculum strategy:
\begin{align*}
(c_\textnormal{targ}^t, c_t) &\gets \argmax_{(c_\textnormal{targ}, c) \in \widehat{\mathcal{C}}_\textnormal{targ} \times \widehat{\mathcal{C}}_\textnormal{unif}} Z_t(c) \cdot Z_t(c_\textnormal{targ}) \cdot \ipp{\psi(c)}{\psi(c_\textnormal{targ})} .
\end{align*}
\looseness-1Next, we replace $V^* (\cdot)$ with $V_\textnormal{max}$, i.e., the maximum possible value that can be achieved for any task in the task space -- this value can typically be obtained for a given domain. Further, when training deep RL agents, allowing some stochasticity in task selection is useful. In particular, the $\argmax$ selection can be problematic in the presence of any approximation errors while computing $V^{\pi_{\theta_t}} (\cdot)$ values. To make the selection more robust, we replace $\argmax$ selection with softmax selection and sample $(c_\textnormal{targ}^t, c_t)$ from the distribution given below:
\begin{align}
& \mathbb{P}\big[(c_\textnormal{targ}^t, c_t) = (c_\textnormal{targ}, c)\big] ~\propto~ \exp \Big(\beta \cdot \frac{V^t (c)}{V_\textnormal{max}} \cdot \big(V_\textnormal{max} - V^t (c)\big) \nonumber \\ & \quad \cdot \frac{V^t (c_\textnormal{targ})}{V_\textnormal{max}} \cdot \big(V_\textnormal{max} - V^t (c_\textnormal{targ})\big) \cdot \ipp{\psi(c)}{\psi(c_\textnormal{targ})}\Big),
\label{eq:cta-scheme-rl-softmax}
\end{align}
\looseness-1where $\beta$ is a hyperparameter and $V^t (\cdot)$ values are obtained from the critic network of the RL agent to estimate $V^{\pi_{\theta_t}} (\cdot)$. Finally, the teacher component samples $(c_\textnormal{targ}^t, c_t)$ from the above distribution and provides the task $c_t$ to the student component -- we refer to this selection strategy as \AlgoOurs{}.

%% file: figs/environments/fig_environments.tex
% !TEX root = main.tex
%%%%%%%%%%%%%%%%%%%%%%%%%%%%%%%%%%%%%
%%%%%%%%%%%%%%%%%%%%%%%%%%%%%%%%%%%%%
%%%%%%%%%%%%%%%%%%%%%%%%%%%%%%%%%%%%%
\begin{figure*}[t!]
\centering
    %%%%%%%%%%%%%%%%%
   %
    %%%%%%%%%%%%%%%%%
    \begin{subfigure}[b]{\textwidth}
    \centering
    {
        \scalebox{0.95}{
    	\setlength\tabcolsep{7pt}
    	\renewcommand{\arraystretch}{1.45}
    	\begin{tabular}{r||c|c|c|c|c}
			 & \envPointMassS{}\textsc{:1T} & \envPointMassS{}\textsc{:2G} & \envSGR{}
                & \textsc{MiniG} 
                & \envBipedalWalker \\
		\toprule
            Reward & binary & binary & binary & binary & dense \\
            Context & $\mathbb{R}^3$ & $\mathbb{R}^3$ & $\mathbb{R}^3$ & $\{0,1\}^8$ & $\mathbb{R}^2$ \\
            State & $\mathbb{R}^{4}$ & $\mathbb{R}^{4}$ & $\mathbb{R}^{4}$ & $[0, 255]^{147}$ & $\mathbb{R}^{24}$ \\
            Action & $\mathbb{R}^2$ & $\mathbb{R}^2$ & $\mathbb{R}^2$ & 7 & $\mathbb{R}^4$ \\ 
            Target Dist. & Single Task & \makecell{Double-Mode \\ Gaussian} & $\mathbb{R}^2$ Plane & Single Task & \makecell{Uniform with \\ trivial tasks} \\
		\bottomrule
        \end{tabular}
        }
        \caption{Complexity of environments} 
        \label{fig:env.1}
    }
    \end{subfigure}%
    %%%%%%%%%%%%%%%%%
    
    \bigskip
    \begin{subfigure}{0.90\textwidth}
    \centering
	{
	\begin{minipage}{0.16\textwidth}
		\includegraphics[height=2.8cm]
            {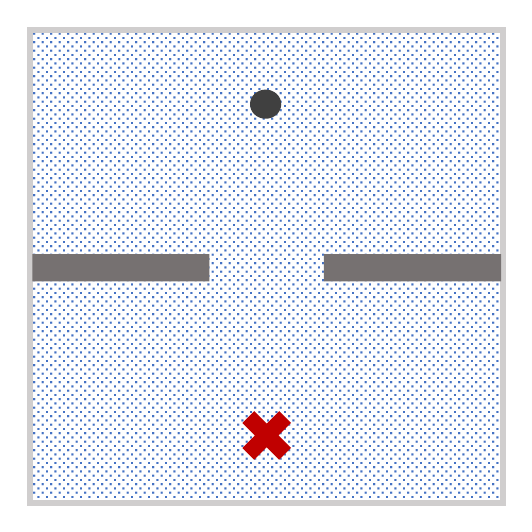}    
        \end{minipage}
	\begin{minipage}{0.24\textwidth}     
            \includegraphics[height=2.8cm]{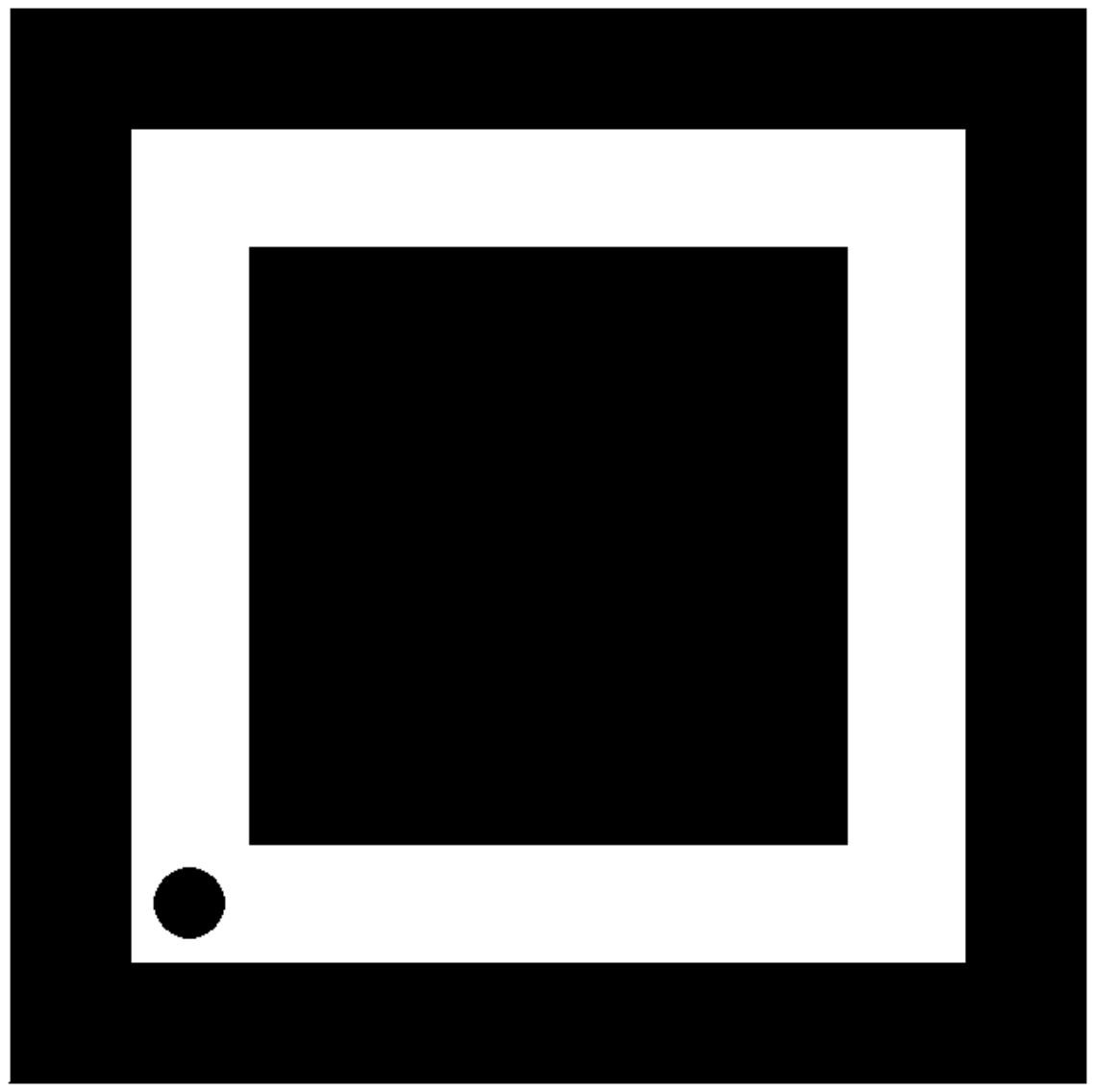}
        \end{minipage}
        \begin{minipage}{0.33\textwidth}     
            \includegraphics[height=2.60cm]{figs/environments/Minigrid.pdf}
        \end{minipage}    
        \begin{minipage}{0.24\textwidth}
		    \includegraphics[height=2.8cm]{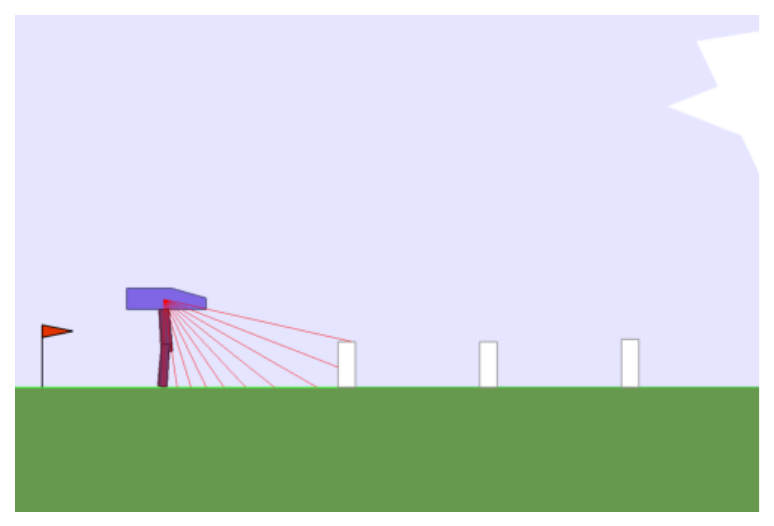}
        \end{minipage} 
        \caption{Illustration of the environments}
	    \label{fig.experiments.environments.illustrations}        
	}
    \end{subfigure}
    %%%%%%%%%%%%%%%%%
    
    %
    % %%%%%%%%%%%%%%%%%% 

    %   
    %%%%%%%%%%%%%%%%%
    %%%%%%%%%%%%%%%%%
	\caption{\textbf{(a)} provides a comprehensive overview of the complexity of the environments based on the reward signals, context space, state space, action space, and target distribution. \textbf{(b)} showcases the environments by providing an illustrative visualization of each environment (from left to right): \envPointMassS{}, \envSGR{}, \textsc{MiniG}, and \envBipedalWalker{}. 
    %Refer to Section~\nameref{sec:exp-envs} for further details.
    }
    \label{fig:env}
    %\vspace{2mm}
\end{figure*}
%%%%%%%%%%%%%%%%%%%%%%%%%%%%%%%%%%%%%

%% file: 4_experiments.tex
% !TEX root =  main.tex
%%%%%%%%%%%%%%%%%%%%%%%%%%%%%%%%%%%%%
%%%%%%%%%%%%%%%%%%%%%%%%%%%%%%%%%%%%%

\section{Experimental Evaluation}
\label{sec:experiments}
In this section, we validate the effectiveness of our curriculum strategy by conducting experiments in environments selected from the state-of-the-art works of~\cite{klink2022curriculum}~and~\cite{romac2021teachmyagent}. We utilize the PPO method from the Stable-Baselines3 library for policy optimization~\cite{schulman2017proximal,stable-baselines3}. The implementation details of different curriculum strategies are provided in appendices of the paper.

\subsection{Environments}
\label{sec:exp-envs}
In our evaluation, we examine four distinct environments detailed in the following paragraphs. These environments are selected to showcase the utility of our curriculum strategy in diverse settings, such as in procedural task generation or with image-based observations, and its effectiveness in handling target distributions with varying characteristics within the context space $\mathcal{C}$. For the first environment, \emph{Point Mass Sparse} (\envPointMassS{}), we consider two settings. In one setting, the target is concentrated on a single context $c \in \mathcal{C}$, a similar setting as analyzed in Section~\ref{sec:curr-discrete-single}. In the second setting, the target distribution exhibits multiple modalities. The second environment, \emph{Sparse Goal Reaching} (\envSGR{}), features target distributions with uniform coverage over specific dimensions of the context space and concentrated on one dimension. The third environment, MiniGrid (\textsc{MiniG}), uniquely features a discrete context space and additionally has an image-based state space. Lastly, a fourth environment, Bipedal Walker Stump tracks (\envBipedalWalker{}), has a uniform target distribution spanning the entirety of the context space. Moreover, it shows the applicability of our technique in the procedural task generation domain. A summary and illustration of these environments are presented in Figure~\ref{fig:env}. 

\textbf{\emph{Point Mass Sparse} (\envPointMassS{}).} Based on the work of~\cite{klink2020selfdeep}, we consider a contextual \envPointMassS{} environment where an agent navigates a point mass through a gate of a given size towards a goal in a two-dimensional space. To heighten the challenge, we replace the original dense reward function with a sparse one, a strategy also considered in~\cite{tzannetos2023proximal}. Specifically, in the \envPointMassS{} environment, the agent operates within a goal-based reward setting where the reward is binary and sparse, i.e., the agent receives a reward of $1$ only upon successfully moving the point mass to the goal position. The parameters governing this environment, such as the gate's position, width, and the ground's friction coefficient, are controlled by a contextual variable $c \in \mathcal{C} \subseteq \mathbb{R}^3$. This variable comprises \emph{C-GatePosition}, \emph{C-GateWidth}, and \emph{C-Friction}. Our experimental section explores two distinct \envPointMassS{} environment settings. In the first setting, denoted as \envPointMassS{}\textsc{:2G}, the target distribution $\mu$ takes the form of a bimodal Gaussian distribution. Here, the means of the contextual variables $\bigl[$\emph{C-GatePosition}, \emph{C-GateWidth}$\bigr]$ are set to $\bigl[-3.9, 0.5\bigr]$ and $\bigl[3.9, 0.5\bigr]$ for the two modes, respectively. In the second setting, \envPointMassS{}\textsc{:1T}, the target distribution $\mu$ is concentrated on a single context $c \in \mathcal{C}$. More precisely, the contextual variables $\bigl[$\emph{C-GatePosition}, \emph{C-GateWidth}, \emph{C-Friction}$\bigr]$ take on the following values: $\bigl[0.9, 0.5, 3.5\bigr]$. To construct our training tasks, we draw $20000$ contexts from $\mathcal{C}$ using a uniform distribution, forming $\widehat{\mathcal{C}}_\textnormal{unif}$. 
The set $\widehat{\mathcal{C}}_\textnormal{targ}$ is created by sampling $400$ contexts from $\mathcal{C}$ according to the target distribution $\mu$. We employ a held-out set sampled from the target $\mu$ for evaluation.

\textbf{\emph{Sparse Goal Reaching} (\envSGR{}).} Based on the work of~\cite{klink2022curriculum}, we consider a sparse-reward, goal-reaching environment in which an agent needs to reach a desired position with high precision. Such environments have previously been studied by~\cite{florensa2018automatic}. Within this environment, the contexts, denoted as $c \in \mathcal{C} \subseteq \mathbb{R}^3$, encode both the desired 2D goal position and the acceptable tolerance for reaching that goal. Our primary objective centers around achieving as many goals as possible with high precision, indicated by a low tolerance threshold. In this regard, the target distribution $\mu$ takes the form of a uniform distribution, but it is restricted to a specific 2D region within $\mathcal{C}$ where the tolerance (\emph{C-Tolerance}) for each context is set at a minimal value of $0.05$. Additionally, the presence of walls within the environment renders many of the tasks specified by  $\mathcal{C}$ infeasible, necessitating the identification of a feasible task subspace. We generate our training tasks by randomly selecting $9900$ contexts from $\mathcal{C}$ using uniform distribution to create $\widehat{\mathcal{C}}_\textnormal{unif}$, and by selecting $100$ contexts according to the target distribution $\mu$ to form $\widehat{\mathcal{C}}_\textnormal{targ}$. For evaluation, we employ a separate held-out set sampled from the target distribution $\mu$.

\textbf{{\emph{MiniGrid}} (MiniG).} We establish the \textsc{MiniG} environment by assembling six diverse Minigrid environments from~\cite{MinigridMiniworld23}: Crossing, Dynamic Obstacles, Four Rooms, Unlock, Unlock Pickup, and Blocked Unlock Pickup. Each environment presents a unique mission, demanding distinct skills such as navigation, goal-reaching, lava avoidance, moving obstacle avoidance, key picking, door unlocking, object picking, and door unblocking. These skills define the discrete context space $\{0,1\}^8$ of \textsc{MiniG}, with each environment requiring a specific subset of skills for a successful resolution. The type Blocked Unlock Pickup is chosen as the target environment due to its inherent difficulty, making it challenging to solve without a curriculum. Additionally, \textsc{MiniG} includes environments like Crossing and Dynamic Obstacles, featuring skills not pertinent to the target mission. The state space comprises observed images of the grid world, and the action space is discrete. The reward is set at $1$ for successful mission completion and $0$ otherwise. For training tasks, we select $1000$ instances from all six environment types. The first three types (Crossing, Dynamic Obstacles, and Four Rooms) collectively contribute to $75\%$ of the training tasks, while the remaining three types, including samples from the target environment, equally constitute the remaining $25\%$.

\textbf{\emph{Bipedal Walker Stump Tracks} (\envBipedalWalker{}).} We conduct additional experiments within the TeachMyAgent benchmark for curriculum techniques, as introduced in~\cite{romac2021teachmyagent}. In this context, we chose a bipedal agent tasked with walking in the Stump Tracks environment, which is an extension of the environment initially proposed in~\cite{portelas19curriculum}. The state space comprises lidar sensors, head position, and joint positions. The action space is continuous, and the goal is to learn a policy that controls the torque of the agent's motors. The walker is rewarded for going forward and penalized for torque usage. An episode lasts $2000$ steps and is terminated if the agent reaches the end of the track or if its head collides with the environment (in which case a reward of $-100$ is received). Within this environment, the contextual variables $c \in \mathcal{C} \subseteq \mathbb{R}^2$ control the height (\emph{C-StumpHeight}) and spacing (\emph{C-StumpSpacing}) of stumps placed along the track for each task. Our experimental setup is equivalent to the bipedal walker stump track environment with mostly trivial tasks, as described in~\cite{romac2021teachmyagent}. In this setup, \emph{C-StumpHeight} is constrained to the range $\bigl[ -3; 3\bigr]$, while \emph{C-StumpSpacing} remains within $\bigl[0; 6\bigr]$. Notably, the environment enforces the clipping of negative values for \emph{C-StumpHeight}, setting them to $0$. Consequently, half of the tasks have a mean stump height of $0$, introducing a significant proportion of trivial tasks ($50\%$). To address the procedural task generation, we randomly draw $1000$ tasks from $\mathcal{C}$ to construct the training task set, denoted as $\widehat{\mathcal{C}}_\textnormal{unif}$. Additionally, every four epochs, we resample $1000$ tasks and update the training set $\widehat{\mathcal{C}}_\textnormal{unif}$. The set $\widehat{\mathcal{C}}_\textnormal{targ}$ is obtained by sampling $500$ tasks from $\mathcal{C}$ according to the target distribution $\mu$, which is uniform in $\mathcal{C}$.

\subsection{Curriculum Strategies Evaluated}
\label{sec:exp-met-currs}

\textbf{Variants of our curriculum strategy.} We consider two curriculum strategies as described next. First, \AlgoOurs{} is based on Eq.~\eqref{eq:cta-scheme-rl-softmax}. Throughout all the experiments, we use the following choice to compute the similarity between $\psi(s)$ and $\psi(c_\textnormal{targ})$: $\textnormal{exp}(-\norm{c-c_\textnormal{targ}}_2)$. Second, \AlgoOursUn{} is a variant of it that does not take into account the target distribution $\mu$ and hence ignores the correlations. Specifically, \AlgoOursUn{} drops the target task-related terms $\textcircled{B}$ and $\textcircled{C}$ derived in Eq.~\eqref{eq:cta-scheme-rl}, and selects the next task according to the following distribution: $\mathbb{P}\big[c_t = c\big] \propto \exp \big(\beta \cdot \frac{V^t (c)}{V_\textnormal{max}} \cdot (V_\textnormal{max} - V^t (c)) \big)$. We note that this strategy is similar to a ZPD-based curriculum strategy proposed in \cite{tzannetos2023proximal} for uniform performance objectives.

\textbf{State-of-the-art baselines.} \textsc{SPDL}~\cite{klink2020selfdeep}, \textsc{CURROT}~\cite{klink2022curriculum}, \textsc{PLR}~\cite{jiang2021prioritized}, and \textsc{Gradient}~\cite{huang2022curriculum} are state-of-the-art curriculum strategies for contextual RL. We adapt the implementation of an improved version of \textsc{SPDL}, presented in~\cite{klink2021probabilistic}, to work with a discrete pool of contextual tasks. \textsc{PLR}~\cite{jiang2021prioritized} was originally designed for procedurally generated content settings, but we have adapted its implementation for the contextual RL setting operating on a discrete pool of tasks.

\textbf{Prototypical baselines.} We consider two prototypical baselines: \textsc{IID} and \textsc{Target}. The \textsc{IID} strategy samples the next task from $\mathcal{C}$ with a uniform distribution, while the \textsc{Target} strategy samples according to the target distribution $\mu$.

%%%
\input{figs/performance/fig_Performance}
%%%

\subsection{Results}

\textbf{Convergence behavior.} As illustrated in Figure~\ref{fig:results_conv}, the RL agents trained using our curriculum strategy, \AlgoOurs{}, perform competitively w.r.t. those trained with state-of-the-art and prototypical baselines. For \envPointMassS{}\textsc{:1T}, in Figure~\ref{fig:fig:results_conv.1}, we observe that \AlgoOurs{} quickly succeeds in the single target task compared to the other techniques. Although \textsc{CURROT} and \textsc{Gradient} converge slower, they finally perform similarly to the proposed technique. The results for \envPointMassS{}\textsc{:2G} are presented in Figure~\ref{fig:fig:results_conv.2}, where we can observe that \AlgoOurs{}, \textsc{CURROT} and \textsc{Gradient} outperform the other strategies. \AlgoOurs{} demonstrates success in handling bimodal target distributions by alternating the selection between the modes of the target distribution. Although it initially has a slower performance than \AlgoOursUn{} and \textsc{CURROT}, it quickly matches and surpasses their performance. Despite \AlgoOursUn{} not explicitly considering the target distribution in its formulation, it progressively selects more challenging contexts and effectively encompasses the tasks from the target distribution in this scenario. In Figure \ref{fig:fig:results_conv.3} for \envSGR{}, \AlgoOurs{} outperforms all the other techniques. \AlgoOurs{} selects tasks that are neither too hard nor too easy for the agent's current policy and are also correlated with the target distribution. \textsc{CURROT} stands out among other strategies due to its ability to gradually choose tasks from the target distribution. Importantly, solely selecting target contexts for training is inadequate, as evidenced by the underperformance of \textsc{Target} compared to all other techniques. Similarly, in Figure~\ref{fig:fig:results_conv.4} for \textsc{MiniG}, \AlgoOurs{} outperforms all the other techniques by a large margin. For \envBipedalWalker{}, Figure~\ref{fig:fig:results_conv.5}, where the target distribution is uniform, \AlgoOurs{} and \AlgoOursUn{} achieve the best performance. Although, \AlgoOursUn{}, by definition, considers a uniform performance objective, \AlgoOurs{} is capable of succesfully handling a uniform target distribution.

%%%
\input{figs/curriculum12092023/fig_Curriculum_trend}
%%%

\textbf{Curriculum plots.}  Figures~\ref{fig:results_curr_PM_GatePosition}~and~\ref{fig:results_curr_PM_GateWidth} display the average distance between the target and the contexts selected from \AlgoOurs{}, \textsc{CURROT}, \textsc{IID}, and \textsc{Target}. We observe that \AlgoOurs{} and \textsc{CURROT} manage to reduce the average context distance below that of \textsc{IID}, indicating that both techniques gradually prioritize tasks that align with the target. However, it is noteworthy that \textsc{CURROT} continues to decrease the context values to reach the target. Whereas \AlgoOurs{}, after succeeding on the target, returns closer to \textsc{IID} sampling.  Figure~\ref{fig:results_curr_PointMass2d} provides a visual representation of the two-dimensional context space for the \envPointMassS{}\textsc{:2G} setting. The curriculum initially starts from larger \emph{C-GateWidth} values and centered \emph{C-GatePosition} values, gradually shifting towards the two modes of the target in the later stages of training. In Figure~\ref{fig:results_curr_SGR_Tolerance}, we display the average \emph{C-Tolerance} of selected tasks in \envSGR{}. Our findings indicate a consistent trend with Figures~\ref{fig:results_curr_PM_GatePosition}~and~\ref{fig:results_curr_PM_GateWidth}. Both \AlgoOurs{} and \textsc{CURROT} reduce the average \emph{C-Tolerance} below that of \textsc{IID}. However, \AlgoOurs{} does not necessarily converge to the target. Conversely, \textsc{CURROT} persists in reducing the context values to attain convergence with the target. In Figure~\ref{fig:results_curr_Bipedal2d}, we depict the two-dimensional context space for the \envBipedalWalker{} setting. Despite the uniformity of the target distribution of contexts, we observe that in the later stages of training, \AlgoOurs{} disregards trivial tasks characterized by \emph{C-StumpHeight} values smaller than $0$. Instead, it focuses on tasks from the remaining task space.

%% file: figs/performance/fig_Performance.tex
% !TEX root =  main.tex
%%%%%%%%%%%%%%%%%%%%%%%%%%%%%%%%%%%%%
%%%%%%%%%%%%%%%%%%%%%%%%%%%%%%%%%%%%%
%%%%%%%%%%%%%%%%%%%%%%%%%%%%%%%%%%%%%
\begin{figure*}[t!]
\centering
    %%%%%%%%%%%%%%%%%
    \begin{subfigure}[b]{.30\textwidth}
    \centering
    {
        \includegraphics[height=4.8cm]{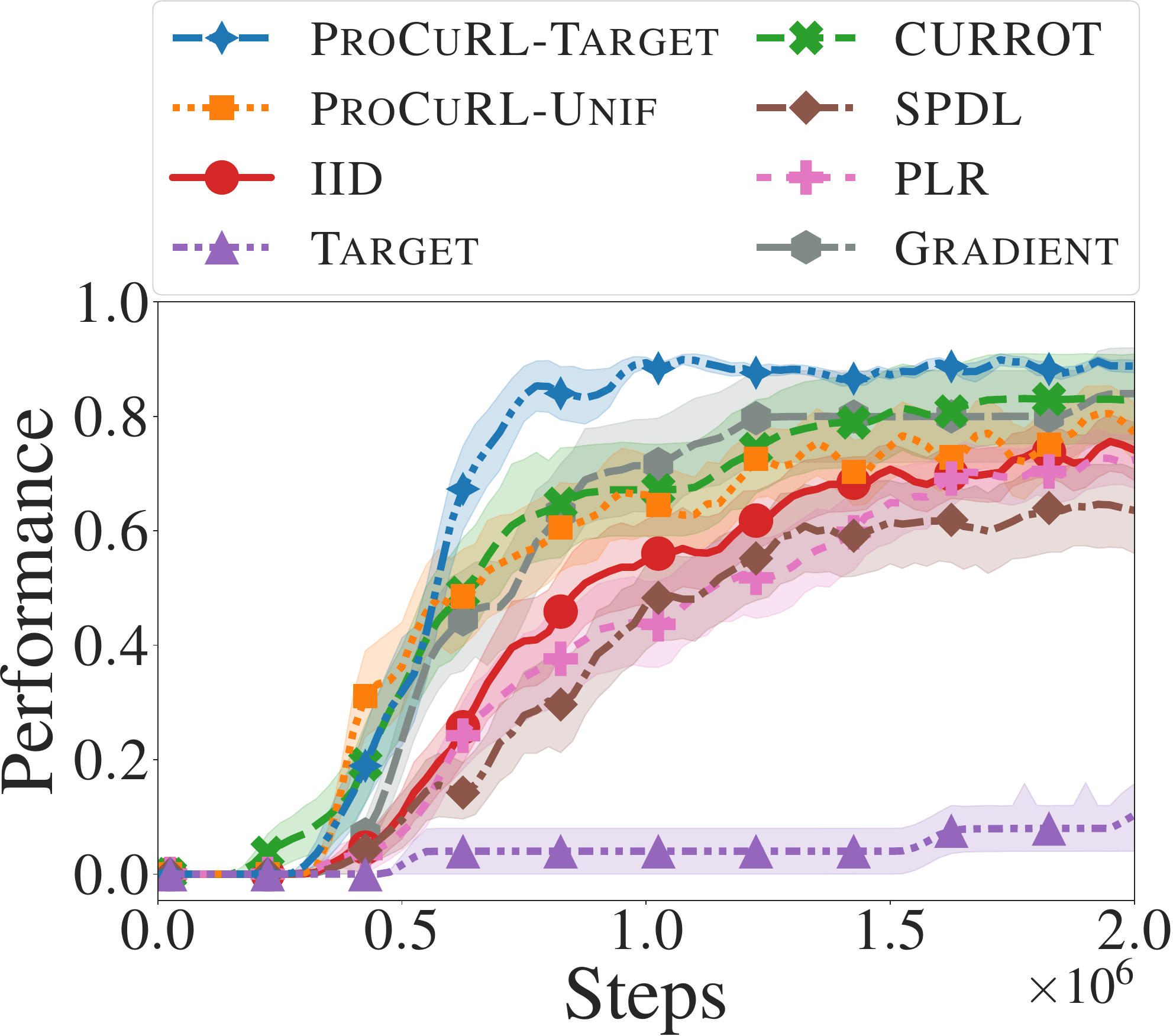}
        \caption{\envPointMassS{}\textsc{:1T}} 
        \label{fig:fig:results_conv.1}
    }
    \end{subfigure}
    %%%%%%%%%%%%%%%%%
    \begin{subfigure}[b]{.30\textwidth}
    \centering
    {
        \includegraphics[height=4.8cm]{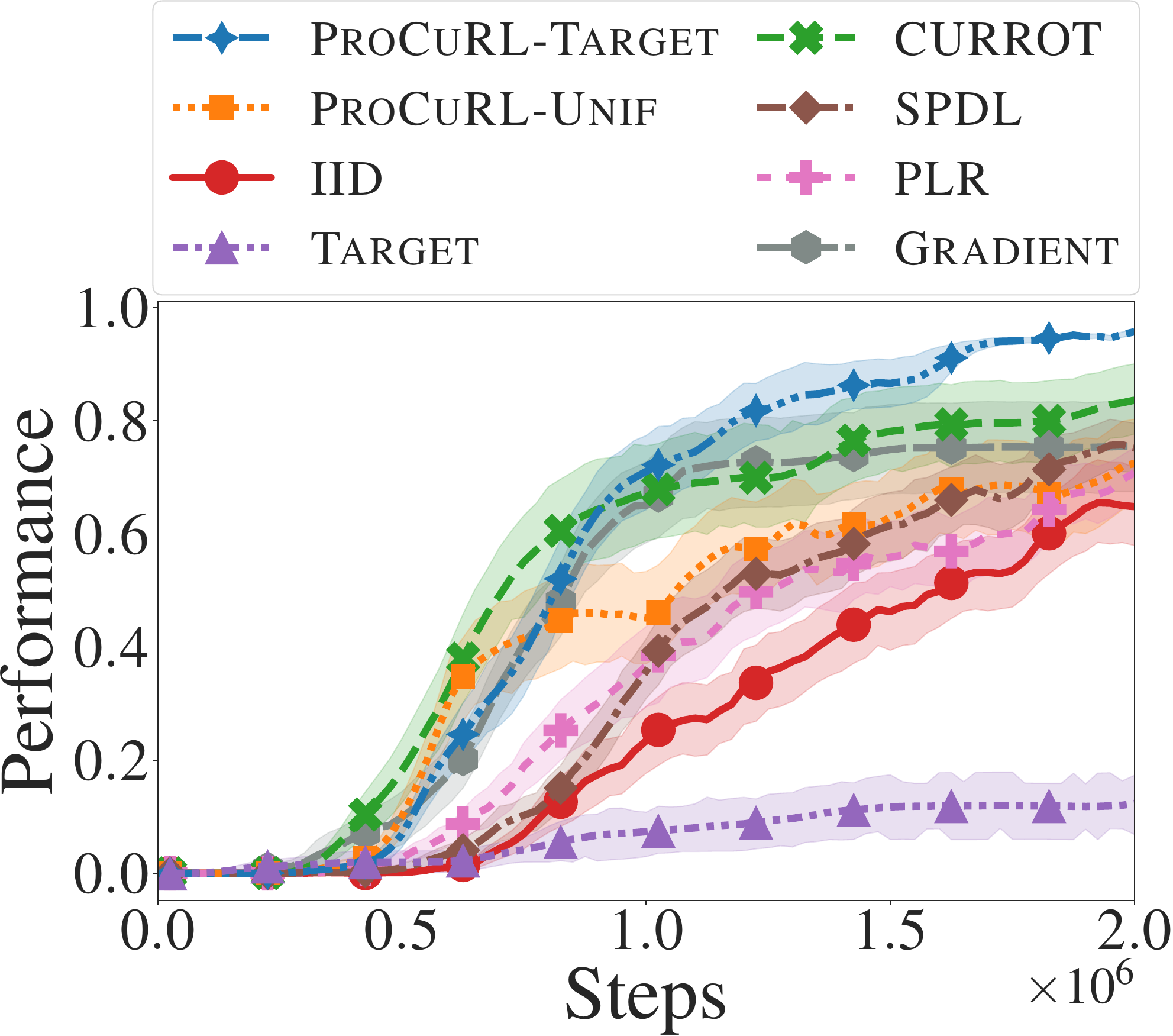}
        \caption{\envPointMassS{}\textsc{:2G}} 
        \label{fig:fig:results_conv.2}
    }
    \end{subfigure}
    % %%%%%%%%%%%%%%%%% 
     %%%%%%%%%%%%%%%%%
    \begin{subfigure}[b]{.30\textwidth}
    \centering
     {
        \includegraphics[height=4.8cm]{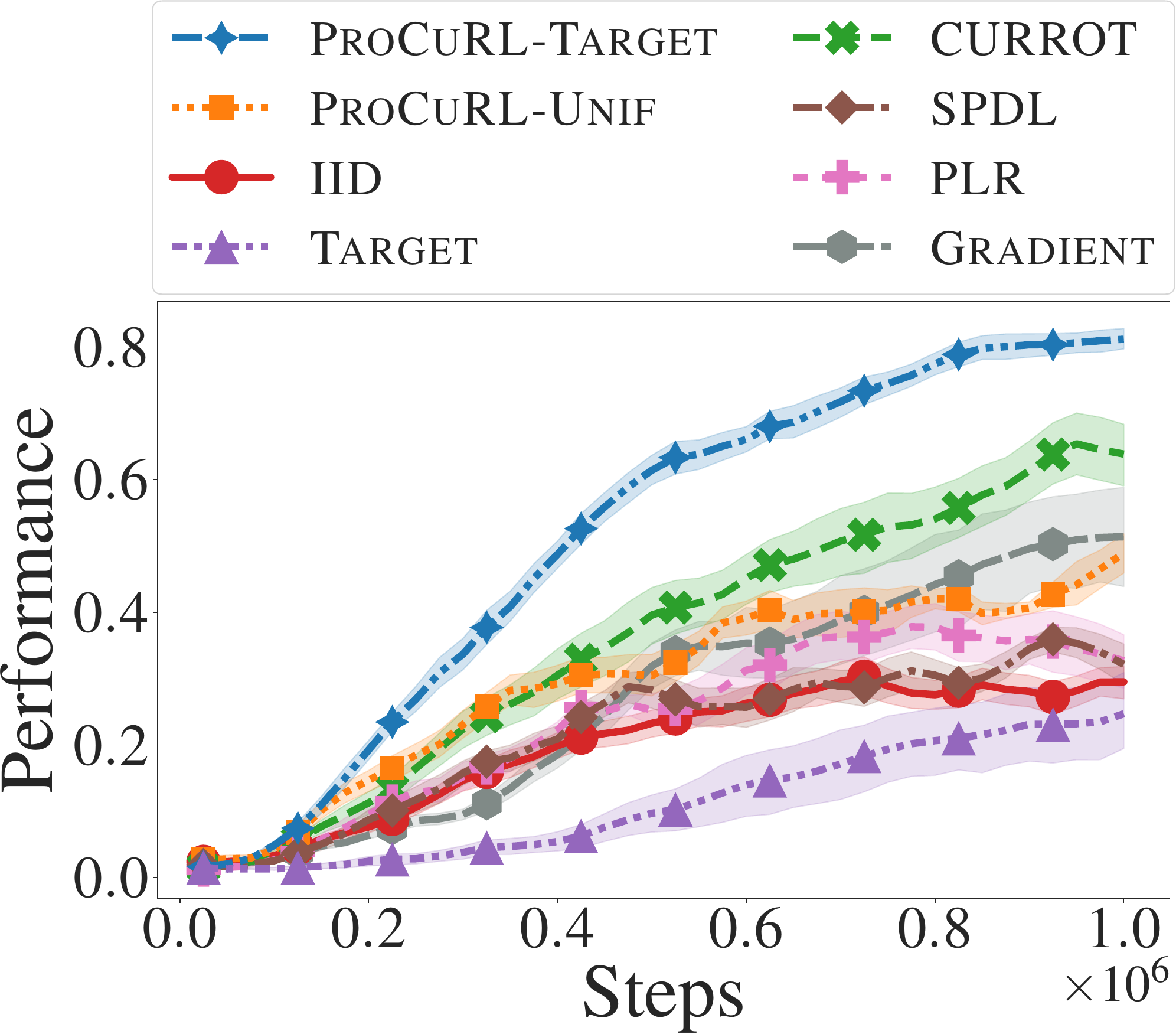}
        \caption{\envSGR} 
        \label{fig:fig:results_conv.3}
    }
    \end{subfigure}
    % %%%%%%%%%%%%%%%%%  

    \bigskip
     %%%%%%%%%%%%%%%%%
    \begin{subfigure}[b]{.35\textwidth}
    \centering
     {
        \includegraphics[height=4.8cm]{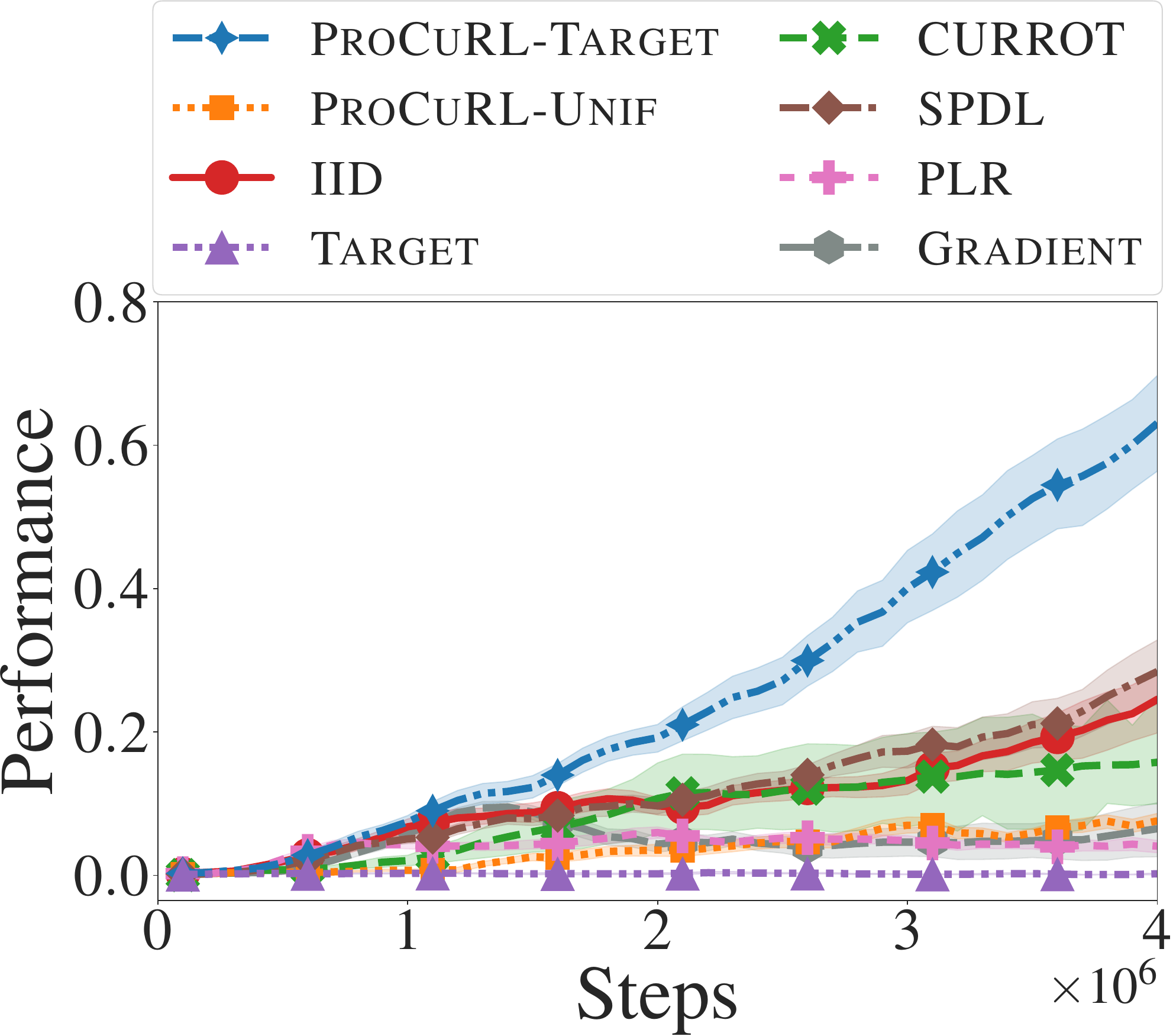}
        \caption{\textsc{MiniG}} 
        \label{fig:fig:results_conv.4}
    }
    \end{subfigure}
    % %%%%%%%%%%%%%%%%%  
     %%%%%%%%%%%%%%%%%
    \begin{subfigure}[b]{.35\textwidth}
    \centering
    {
        \includegraphics[height=4.8cm]{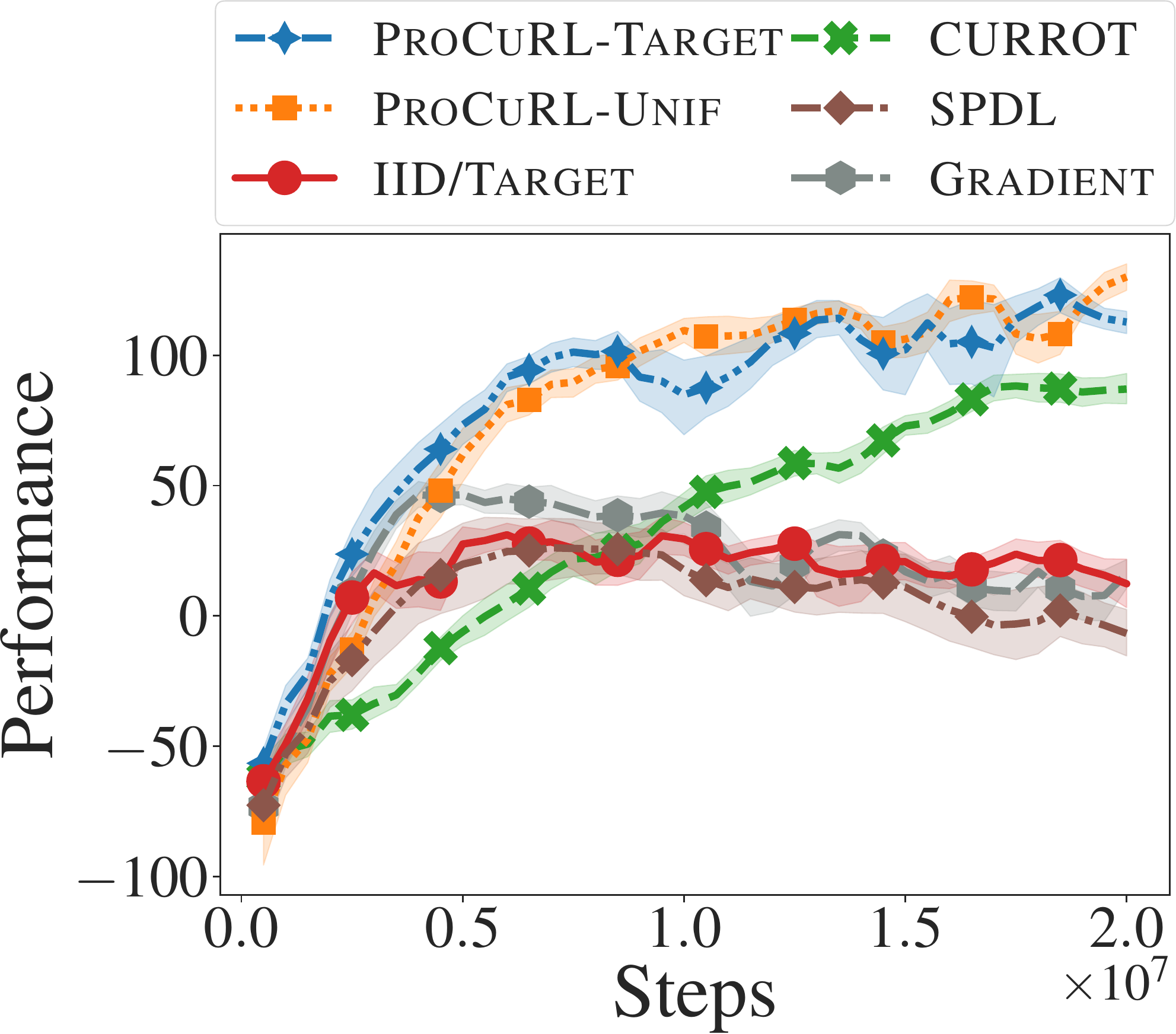}
        \caption{\envBipedalWalker{}} 
        \label{fig:fig:results_conv.5}
    }
    \end{subfigure}
    % %%%%%%%%%%%%%%%%%  
    %
    % %%%%%%%%%%%%%%%%%        
    %%%%%%%%%%%%%%%%%
    %%%%%%%%%%%%%%%%%
	\caption{\looseness-1Performance comparison of RL agents trained using different curriculum strategies.
    %outlined in Section~\ref{sec:exp-met-currs}. 
     The performance is measured as the mean return ($\pm 1$ standard error) on the test pool of tasks. The results are averaged over  $25$ random seeds for \envPointMassS{}\textsc{:1T}, $25$ random seeds for \envPointMassS{}\textsc{:2G}, $10$ random seeds for \envSGR{}, $20$ random seeds for \textsc{MiniG}, and $10$ random seeds for \envBipedalWalker{}. The plots are smoothed across $2$ evaluation snapshots that occur over $25000$ training steps.}
    \label{fig:results_conv}
    %\vspace{2mm}
    %\vspace{-5mm}
\end{figure*}
%%%%%%%%%%%%%%%%%%%%%%%%%%%%%%%%%%%%%

%% file: figs/curriculum12092023/fig_Curriculum_trend.tex
% !TEX root =  main.tex
%%%%%%%%%%%%%%%%%%%%%%%%%%%%%%%%%%%
%%%%%%%%%%%%%%%%%%%%%%%%%%%%%%%%%%%
%%%%%%%%%%%%%%%%%%%%%%%%%%%%%%%%%%%%%
\begin{figure*}[t!]

\captionsetup[subfigure]{%
  justification = RaggedRight, % Or justified
  format=hang}

    \centering
    \begin{subfigure}[b]{.30\textwidth}
    %\begin{subfigure}[b]{.18\textwidth}
    \centering
    {
        \includegraphics[height=4.8cm]{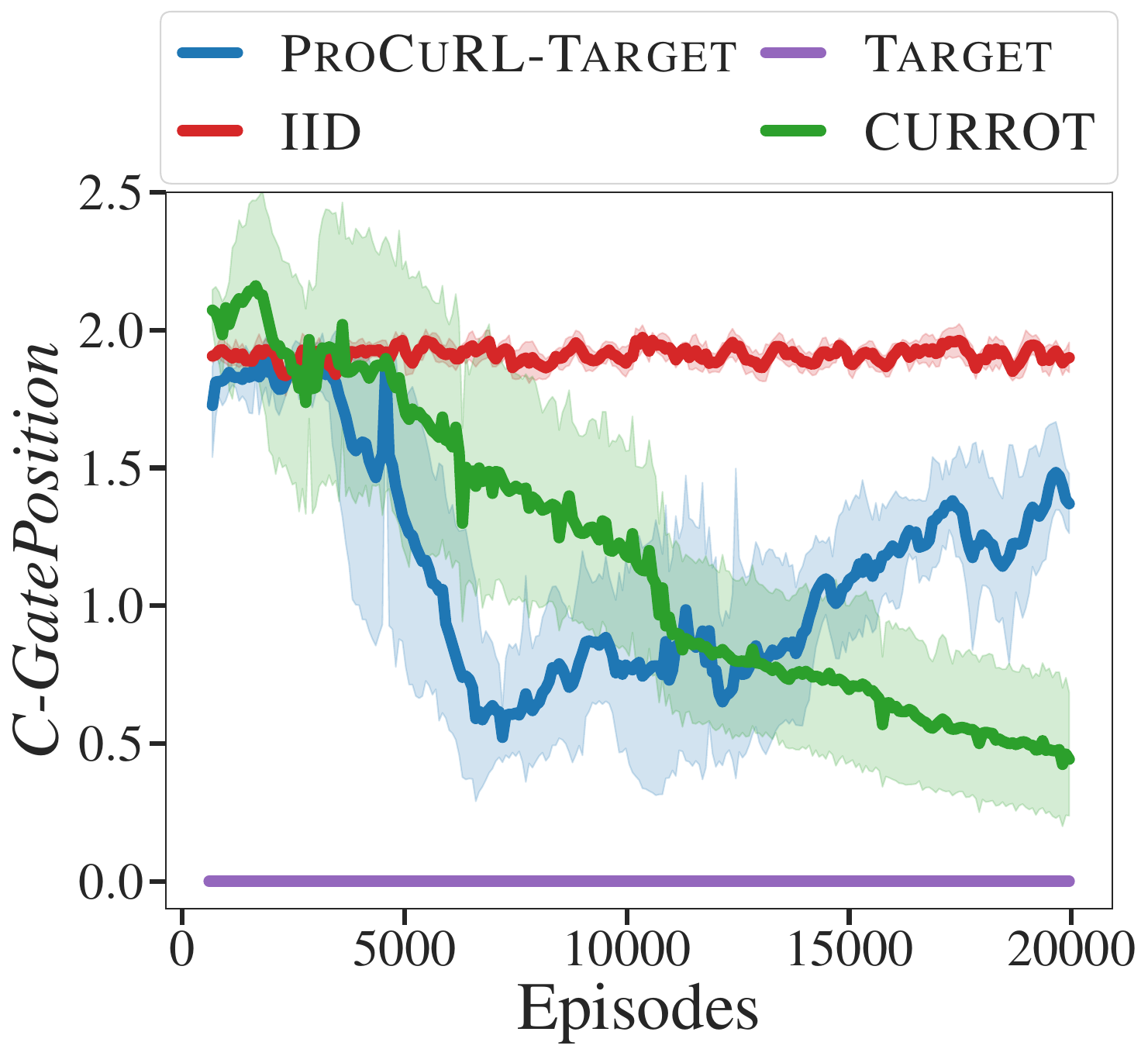}
        \caption{\textsc{PM-s:2G}: \emph{C-GatePosition}} 
        \label{fig:results_curr_PM_GatePosition}
    }
    \end{subfigure}
    %%%%%%%%%%%%%%%%%    
    \begin{subfigure}[b]{.30\textwidth}
    %\begin{subfigure}[b]{.18\textwidth}
    \centering
    {
        \includegraphics[height=4.8cm]{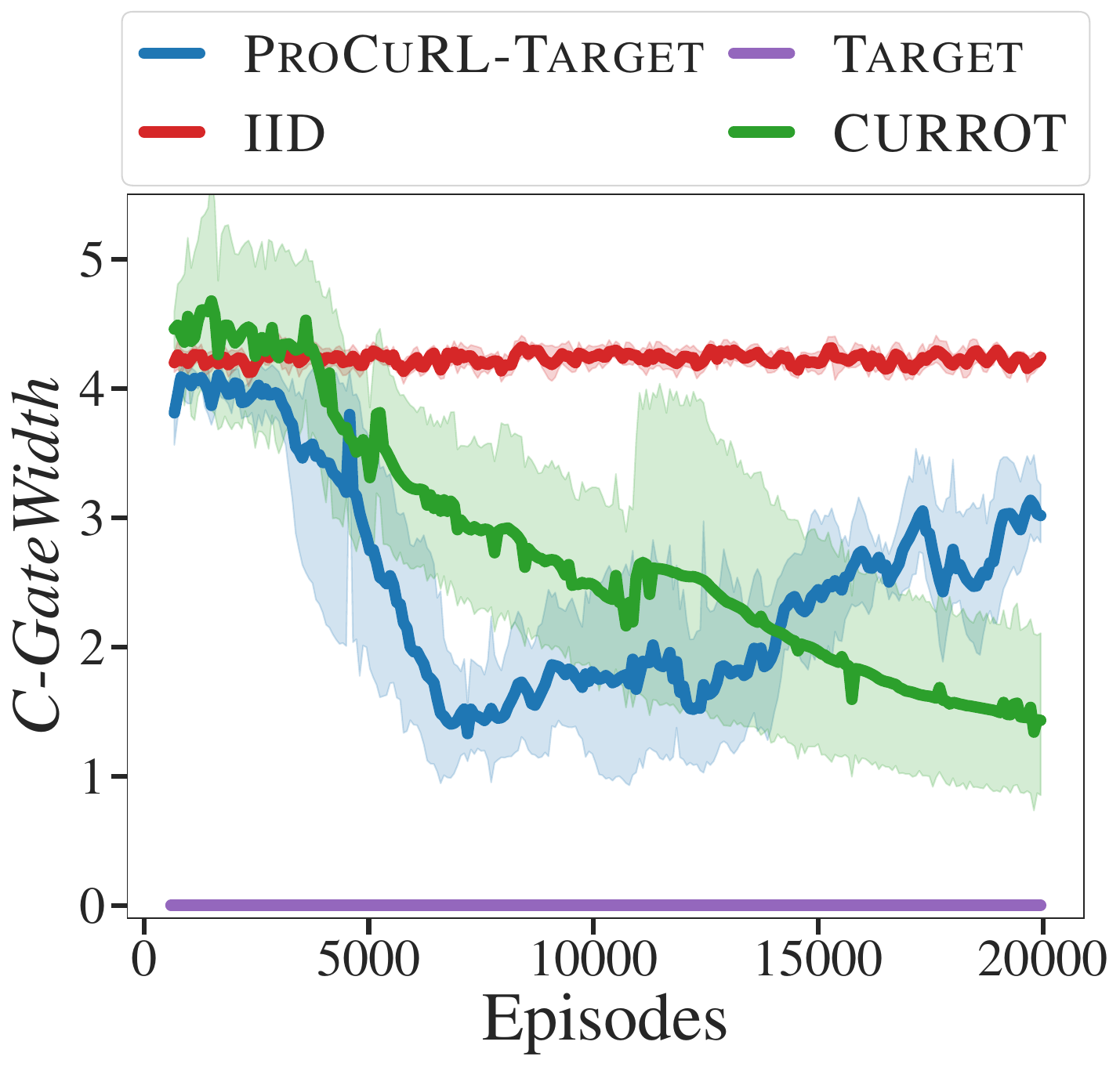}
        \caption{\textsc{PM-s:2G}: \emph{C-GateWidth}} 
        \label{fig:results_curr_PM_GateWidth}
    }
    \end{subfigure}
    %\\
    %%%%%%%%%%%%%%%%%
    %\begin{subfigure}[b]{.18\textwidth}
    \begin{subfigure}[b]{.30\textwidth}
    \centering
    {
        \includegraphics[height=4.8cm]{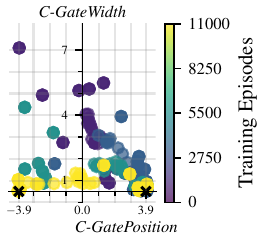}
        \caption{\textsc{PM-s:2G}: 2D context} 
        \label{fig:results_curr_PointMass2d}
    }
    \end{subfigure}
    %%%%%%%%%%%%%%%%%
    
    \bigskip
    %%%%%%%%%%%%%%%%%
    \begin{subfigure}[b]{0.35\textwidth}
    \centering
    {
        \includegraphics[height=4.8cm]{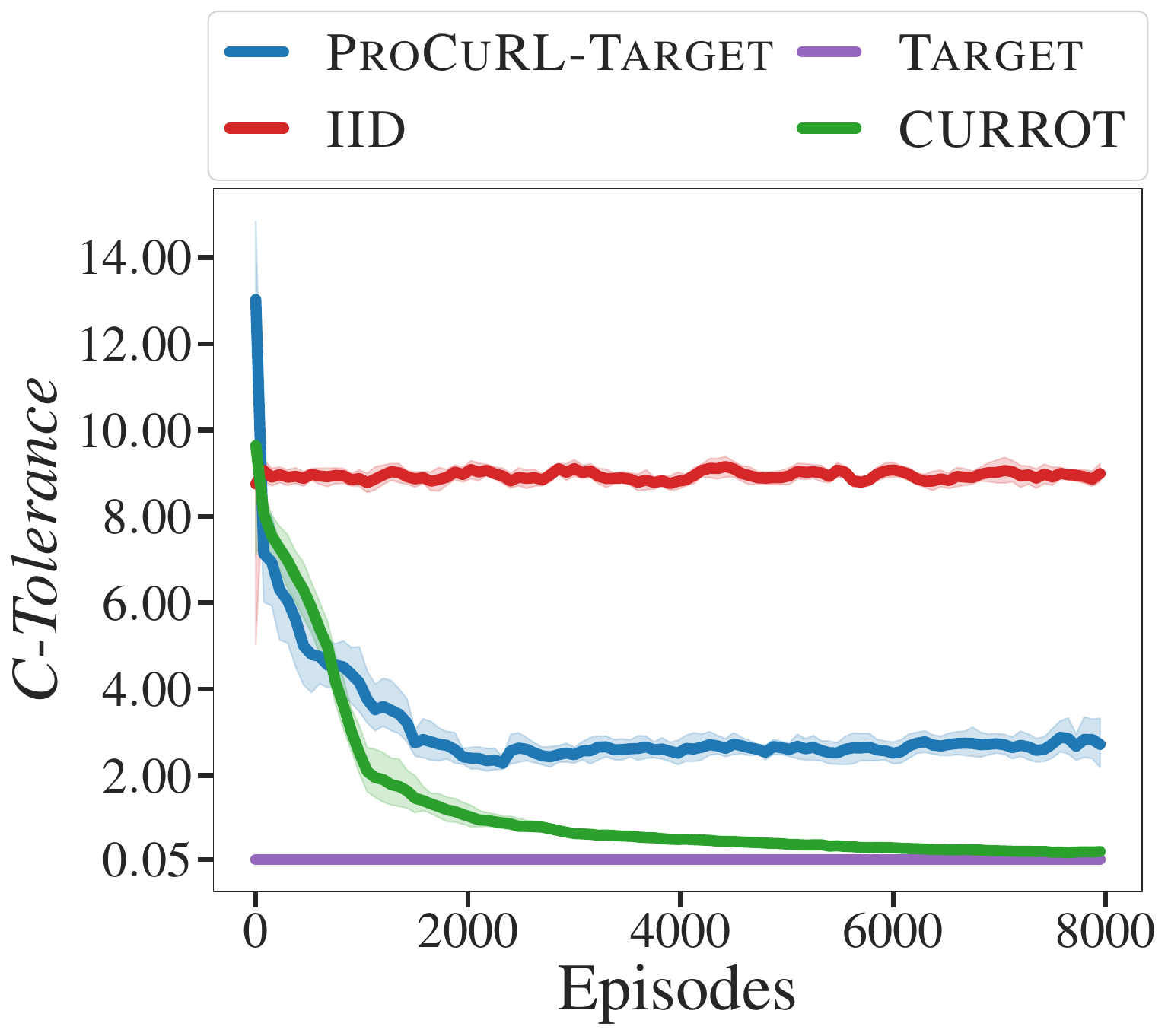}
        \caption{\footnotesize{\envSGR{}:} \emph{C-Tolerance}} 
        \label{fig:results_curr_SGR_Tolerance}
    }
    \end{subfigure}
    %%%%%%%%%%%%%%%%%
    \begin{subfigure}[b]{.35\textwidth}
    \centering
    {
        \includegraphics[height=4.8cm]{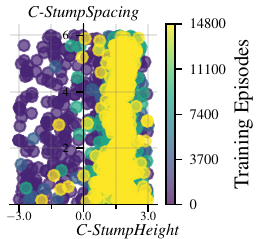}
        \caption{\textsc{BW}: 2D context} 
        \label{fig:results_curr_Bipedal2d}
    }
    \end{subfigure}
    \caption{\textbf{(a-b)} present the average distance between the selected contexts \emph{C-GatePosition} and \emph{C-GateWidth} and the target distribution for \envPointMassS{}\textsc{:2G}. \textbf{(c)} presents the two-dimensional context space of \envPointMassS{}\textsc{:2G}. The target distribution is depicted as a black \textbf{x} and encodes the two gates with \emph{C-GateWidth} $=0.5$ at \emph{C-GatePosition} $=\{-3.9, 3.9\}$.  Each colored dot represents the context/task selected by \AlgoOurs{} during training, where brighter colors indicate later training stages. \textbf{(d)} presents the average \emph{C-Tolerance} of the selected tasks during different curriculum strategies for \envSGR{}. \textbf{(e)} presents the two-dimensional context space of \envBipedalWalker{}. The target distribution is uniform. Each colored dot represents the context/task selected by \AlgoOurs{} during training, where brighter colors indicate later training stages.}
    \label{fig:results_curr_context_trends}
    %\vspace{2mm}
\end{figure*}
%%%%%%%%%%%%%%%%%%%%%%%%%%%%%%%%%%%%%

%% file: 5_conclusion.tex
% !TEX root =  main.tex
%%%%%%%%%%%%%%%%%%%%%%%%%%%%%%%%%%%%%
%%%%%%%%%%%%%%%%%%%%%%%%%%%%%%%%%%%%%

\section{Concluding Discussions}
\label{sec:conclusions}

We proposed a novel curriculum strategy that strikes a balance between selecting tasks that are neither too hard nor too easy for the agent while also progressing the agent's learning toward the target distribution by utilizing task correlation. We mathematically derived our curriculum strategy through an analysis of a specific learning scenario and demonstrated its effectiveness in various environments through empirical evaluations. Here, we discuss a few limitations of our work and outline a plan on how to address them in future work. First,  it would be interesting to extend our curriculum strategy to high-dimensional context spaces in sparse reward environments. However, sampling new tasks in such environments poses a significant challenge due to the estimation of the value of all tasks in the discrete sets $\widehat{\mathcal{C}}_\textnormal{unif}$ and $\widehat{\mathcal{C}}_\textnormal{targ}$. Second, while our curriculum strategy uses a simple distance measure to capture task correlation, it would be worthwhile to investigate the effects of employing different distance metrics over the context space on curriculum design. 

%% file: 9_appendix.tex
% !TEX root =  main.tex
%%%%%%%%%%%%%%%%%%%%%%%%%%%%%%%%%%%%%
%%%%%%%%%%%%%%%%%%%%%%%%%%%%%%%%%%%%%
\appendix
{
    \allowdisplaybreaks
    \input{9.0_appendix_table-of-contents}

    \input{9.1_appendix_discussion}

    \input{9.2_appendix_proof}

    \input{9.3_appendix_experiments}

}

%% file: 9.0_appendix_table-of-contents.tex
% !TEX root =  main.tex
%%%%%%%%%%%%%%%%%%%%%%%%%%%%%%%%%%%%%
%%%%%%%%%%%%%%%%%%%%%%%%%%%%%%%%%%%%%
\section{Table of Contents}
\label{sec-app:toc}

In this section, we briefly describe the content provided in the paper's appendices. 
\begin{itemize}[leftmargin=*]
    \item Appendix~\ref{sec-app:discussion} provides a discussion of the broader impact of our work and compute resources used.
    \item Appendix~\ref{sec-app:theoretical-justification} provides the proofs for Theorem~\ref{thm:curriculum-general} and Proposition~\ref{prop:prox-analysis-reinforce-learner}. (Section~\ref{sec:curr-discrete-single})
    \item Appendix~\ref{sec-app:experiments} provides additional details for experimental evaluation. (Section~\ref{sec:experiments})
\end{itemize}

%% file: 9.1_appendix_discussion.tex
% !TEX root =  main.tex
%%%%%%%%%%%%%%%%%%%%%%%%%%%%%%%%%%%%%
%%%%%%%%%%%%%%%%%%%%%%%%%%%%%%%%%%%%%
\section{Discussions}\label{sec-app:discussion}

\textbf{Broader impact.} This work presents a novel curriculum strategy for contextual multi-task settings where the agent's final performance is measured w.r.t. a target distribution over the context space. Given the algorithmic and empirical nature of our work applied to learning agents, we do not foresee any direct negative societal impacts of our work in the present form. 

\noindent\textbf{Compute resources.} The experiments for \envSGR{}, \envPointMassS{} and \textsc{MiniG} were conducted on a cluster of machines equipped with Intel Xeon Gold 6134M CPUs running at a frequency of 3.20GHz. The experiments for \textsc{BipedalWalker} were conducted on machines equipped with Tesla V100 GPUs.

%% file: 9.2_appendix_proof.tex
% !TEX root =  main.tex
%%%%%%%%%%%%%%%%%%%%%%%%%%%%%%%%%%%%%
%%%%%%%%%%%%%%%%%%%%%%%%%%%%%%%%%%%%%

\section{Justifications for the Curriculum Strategy -- Proofs}
\label{sec-app:theoretical-justification}

\subsection{Proof of Theorem~\ref{thm:curriculum-general}}
\label{sec-app:general-setting}

\begin{proof}
For a task space $\mathcal{C}$ of sufficient complexity, at any given step $t$, the task $c_t$ selected by the curriculum strategy outlined in Eq.~\eqref{eq:curr-gradient-form} adheres to the following policy improvement property: 
\begin{align*}
\Expect{(V^{\pi_{\theta_{t+1}}} (c_\textnormal{targ}) - V^{\pi_{\theta_t}} (c_\textnormal{targ})) \mid \theta_t} ~=~& \beta_t \cdot (V^* (c_\textnormal{targ}) - V^{\pi_{\theta_t}} (c_\textnormal{targ})) - \delta_t ,
\end{align*}
where $\beta_t \in (0,1)$ and $\delta_t \geq 0$. Given that $\Expect{(V^{\pi_{\theta_{t+1}}} (c_\textnormal{targ}) - V^{\pi_{\theta_t}} (c_\textnormal{targ})) \mid \theta_t} \approx \eta_t \cdot \ipp{g_t(c_t)}{g_t(c_\textnormal{targ})}$, for sufficiently complex $\mathcal{C}$, it becomes feasible for the task $c_t$ maximizing $\ipp{g_t(c)}{g_t(c_\textnormal{targ})}$ to fulfill the aforementioned policy improvement property.

Now, let's consider the following expression:
\begin{align*}
\Expect{(V^{*} (c_\textnormal{targ}) - V^{\pi_{\theta_{t+1}}} (c_\textnormal{targ})) \mid \theta_t} ~=~& (V^{*} (c_\textnormal{targ}) - V^{\pi_{\theta_t}} (c_\textnormal{targ})) - \Expect{(V^{\pi_{\theta_{t+1}}} (c_\textnormal{targ}) - V^{\pi_{\theta_t}} (c_\textnormal{targ})) \mid \theta_t} \\
~=~& (V^{*} (c_\textnormal{targ}) - V^{\pi_{\theta_t}} (c_\textnormal{targ})) - \beta_t \cdot (V^* (c_\textnormal{targ}) - V^{\pi_{\theta_t}} (c_\textnormal{targ})) + \delta_t \\
~=~& (1 - \beta_t) \cdot (V^{*} (c_\textnormal{targ}) - V^{\pi_{\theta_t}} (c_\textnormal{targ})) + \delta_t \\
~\leq~& (1 - \beta) \cdot (V^{*} (c_\textnormal{targ}) - V^{\pi_{\theta_t}} (c_\textnormal{targ})) + \delta , 
\end{align*}
where $\beta = \min_t \beta_t$ and $\delta = \max_t \delta_t$. By defining $f(\theta) := (V^{*} (c_\textnormal{targ}) - V^{\pi_{\theta}} (c_\textnormal{targ}))$, we can express the above inequality as follows:
\[
\Expect{f(\theta_{t+1}) | \theta_t} ~\leq~ (1-\beta) \cdot f(\theta_t) + \delta , \quad \forall t . 
\]
Subsequently, using the total expectation result, we derive:
\begin{align*}
\Expect{f(\theta_{t+1}) | \theta_{t-1}} ~=~& \Expect{\Expect{f(\theta_{t+1}) | \theta_t} | \theta_{t-1}} \\
~\leq~& (1-\beta) \cdot \Expect{f(\theta_{t}) | \theta_{t-1}} + \delta \\
~\leq~& (1-\beta)^2 \cdot f(\theta_{t-1}) + \delta \cdot \bss{1 + (1-\beta)} 
\end{align*}
By repeating the above steps, we obtain:
\begin{align*}
\Expect{f(\theta_{t+1}) | \theta_{0}} ~\leq~& (1-\beta)^{t+1} \cdot f(\theta_{0}) + \delta \cdot \sum_{i=0}^t{(1-\beta)^i} \\
~\leq~& (1-\beta)^{t+1} \cdot f(\theta_{0}) + \frac{\delta}{\beta} \\
~\leq~& \frac{\epsilon}{2} + \frac{\epsilon}{2} ~=~ \epsilon ,
\end{align*}
for $t+1 = \brr{\log \frac{1}{1-\beta}}^{-1} \log\frac{2 f(\theta_{0})}{\epsilon} = \mathcal{O}\brr{\log \frac{1}{\epsilon}}$ and $\delta \leq \frac{\epsilon \cdot \beta}{2}$. 
\end{proof}

\subsection{Proof of Proposition~\ref{prop:prox-analysis-reinforce-learner}}
\label{sec-app:contextual-bandit}

\begin{proof}
For the contextual bandit setting described above, we simplify the gradient $g_t(c)$ as follows:
\begin{align*}
g_t(c) ~=~& \bss{\nabla_\theta V^{\pi_{\theta}} (c)}_{\theta = \theta_t} \\
~=~& \Expectover{a \sim \pi_{\theta_t}(\cdot | s, c)}{R_c(s, a) \cdot \bss{\nabla_\theta \log \pi_\theta (a|s, c)}_{\theta = \theta_t}} \\
~=~& \pi_{\theta_t}(a_c^\textnormal{opt} | s, c) \cdot R_c(s, a_c^\textnormal{opt}) \cdot \bss{\nabla_\theta \log \pi_\theta (a_c^\textnormal{opt}|s, c)}_{\theta = \theta_t} \\
~=~& \pi_{\theta_t}(a_c^\textnormal{opt} | s, c) \cdot R_c(s, a_c^\textnormal{opt}) \cdot \brr{\phi(s, c, a_c^\textnormal{opt}) - \Expectover{a' \sim \pi_{\theta_t}(\cdot | s, c)}{\phi(s, c, a')}} \\
~=~& \pi_{\theta_t}(a_c^\textnormal{opt} | s, c) \cdot R_c(s, a_c^\textnormal{opt}) \cdot (1 - \pi_{\theta_t}(a_c^\textnormal{opt} | s, c)) \cdot (\phi(s, c, a_c^\textnormal{opt}) - \phi(s, c, a_c^\textnormal{non})) \\
~=~& \pi_{\theta_t}(a_c^\textnormal{opt} | s, c) \cdot R_c(s, a_c^\textnormal{opt}) \cdot (1 - \pi_{\theta_t}(a_c^\textnormal{opt} | s, c)) \cdot \psi(c) \\
~=~& \frac{\pi_{\theta_t}(a_c^\textnormal{opt} | s, c) \cdot R_c(s, a_c^\textnormal{opt})}{R_c(s, a_c^\textnormal{opt})} \cdot (R_c(s, a_c^\textnormal{opt}) - \pi_{\theta_t}(a_c^\textnormal{opt} | s, c) \cdot R_c(s, a_c^\textnormal{opt})) \cdot \psi(c) \\
~=~& \frac{V^{\pi_{\theta_t}} (c)}{V^* (c)} \cdot \big(V^* (c) - V^{\pi_{\theta_t}} (c)\big) \cdot \psi(c) ,
\end{align*}
where we used the facts that $V^{\pi_{\theta_t}} (c) = \pi_{\theta_t}(a_c^\textnormal{opt} | s, c) \cdot R_c(s, a_c^\textnormal{opt})$ and $V^{*} (c) = R_c(s, a_c^\textnormal{opt})$. Based on the above simplification of $g_t(c)$, we have:
\begin{align*}
\ipp{g_t(c)}{g_t(c_\textnormal{targ})} ~=~& \frac{V^{\pi_{\theta_t}} (c)}{V^* (c)} \cdot \big(V^* (c) - V^{\pi_{\theta_t}} (c)\big) \cdot \frac{V^{\pi_{\theta_t}} (c_\textnormal{targ})}{V^* (c_\textnormal{targ})} \cdot \big(V^* (c_\textnormal{targ}) - V^{\pi_{\theta_t}} (c_\textnormal{targ})\big) \cdot \ipp{\psi(c)}{\psi(c_\textnormal{targ})} .
\end{align*}
\end{proof}

%% file: 9.3_appendix_experiments.tex
% !TEX root = main.tex
%%%%%%%%%%%%%%%%%%%%%%%%%%%%%%%%%%%%%%%%%%%%%%%%%%%%%%%%
%%%%%%%%%%%%%%%%%%%%%%%%%%%%%%%%%%%%%%%%%%%%%%%%%%%%%%%%
\section{Experimental Evaluation -- Additional Details}
\label{sec-app:experiments}

\subsection{Environments}\label{sec-app:exp-envs}

\textbf{Point Mass Sparse (\envPointMassS{}~\cite{klink2020selfdeep}).} In this environment, the state consists of the position and velocity of the point-mass, denoted as $s = [x \, \dot{x} \, y \, \dot{y}]$. The action corresponds to the force applied to the point-mass in a 2D space, represented as $a = [F_x \, F_y]$.
The contextual variable $c = [c_1 \, c_2 \, c_3] \in \mathcal{C} \subseteq \mathbb{R}^3$ comprises the following elements: \emph{C-GatePosition}, \emph{C-GateWidth}, and \emph{C-Friction}. The bounds for each contextual variable are $[-4, 4]$ for \emph{C-GatePosition}, $[0.5, 8]$ for \emph{C-GateWidth}, and $[0, 4]$ for \emph{C-Friction}.
At the beginning of each episode, the agent's initial state is set to $s_0 = [0 \, 0 \, 3 \, 0]$, and the objective is to approach the goal located at position $g = [x \, y] = [0 \, -3]$ with sufficient proximity. If the agent collides with a wall or if the episode exceeds 100 steps, the episode is terminated, and the agent receives a reward of $0$. On the other hand, if the agent reaches the goal within a predefined threshold, specifically when $\norm{g - [x \, y]}_2 < 0.30$, the episode is considered successful, and the agent receives a reward of $1$.
The target distribution $\mu$ is represented by a bimodal Gaussian distribution, with the means of the contexts $[\emph{C-GatePosition} \, \emph{C-GateWidth}]$ set as $[-3.9 \, 0.5]$ and $[3.9 \, 0.5]$ for the two modes, respectively. This choice of target distribution presents a challenging scenario, as it includes contexts where the gate's position is in proximity to the edges of the environment and the gate's width is relatively small.
In this environment, we employ Proximal Policy Optimization (PPO) with 5120 steps per policy update. The batch size used for each update is set to 128, and an entropy coefficient of 0.01 is applied. The MLP policy consists of a shared layer with 64 units, followed by a second separate layer with 64 units for the policy network, and an additional 64 units for the value network. All the remaining parameters of PPO adopt the default settings of Stable Baselines~3~\cite{schulman2017proximal,stable-baselines3}. Furthermore, all the hyperparameters remain consistent across all the curriculum strategies evaluated.

\textbf{Sparse Goal Reaching (\envSGR{}~\cite{klink2022curriculum}).} In this environment, the state consists of the agent's position, denoted as $s = [x \, y]$. The action corresponds to the agent's displacement in a 2D space, represented as $a = [d_x \, d_y]$.
The contextual variable $c = [c_1 \, c_2 \, c_3] \in \mathcal{C} \subseteq \mathbb{R}^3$ comprises the following elements: \emph{C-GoalPositionX}, \emph{C-GoalPositionY}, and \emph{C-Tolerance}. The bounds for each contextual variable are $[-9, 9]$ for \emph{C-GoalPositionX}, $[-9, 9]$ for \emph{C-GoalPositionY}, and $[0.05, 18]$ for \emph{C-Tolerance}.
The reward in this environment is sparse, meaning the agent receives a reward of $1$ only when it reaches the goal. An episode is considered successful if the distance between the agent's position and the goal is below the tolerance, i.e., $\norm{s - [c_1 , c_2]}_2 \leq c_3$. If the agent exceeds the limit of $200$ steps per episode before reaching the goal, the episode terminates with a reward of $0$.
The presence of walls in the environment creates situations where certain combinations of contexts $[c_1 \, c_2 \, c_3]$ are unsolvable by the agent, as it is unable to get close enough to the goal to satisfy the tolerance condition. This suggests that a successful curriculum technique should also be able to identify the feasible subspace of contexts to accelerate the training process.
The target context distribution consists of tasks that are uniformly distributed w.r.t. the contexts $c_1$ (\emph{C-GoalPositionX}) and $c_2$ (\emph{C-GoalPositionY}). However, it is concentrated in the minimal \emph{C-Tolerance}, where $c_3$ is set to $0.05$. In other words, the target distribution comprises only high-precision tasks.
In this environment, we employ Proximal Policy Optimization (PPO) with 5120 steps per policy update and a batch size of 256. The MLP policy consists of a shared layer with 64 units, followed by a second separate layer with 32 units for the policy network, and an additional 32 units for the value network. All the remaining parameters of PPO adopt the default settings of Stable Baselines~3~\cite{schulman2017proximal,stable-baselines3}. Furthermore, all the hyperparameters remain consistent across all the curriculum strategies evaluated.

\textbf{MiniGrid (\textsc{MiniG}~\cite{MinigridMiniworld23}).} In this environment, the state consists of a $7\times7\times3$ image of the grid world with an additional bit for the agent's direction. The action space is discrete and consists of the following actions: left, right, forward, pickup, drop, toggle, done. The utilized actions can be different depending on the type of Minigrid environment. The context $c$ is discrete and represents the set of $8$ skills necessary for the agent to solve the required Minigrid mission. The context space can be represented as $\{0,1\}^8$, where the necessary skills are represented with $1$ and the unnecessary with $0$. Each environment can have a different mission, and \textsc{MiniG} contains $6$ different mission types. As a target, we select the type Blocked Unlock Pickup due to its inherent difficulty, making it challenging to solve without a curriculum. This environment requires the skills of navigation, key picking, door unlocking, object picking, and door unblocking. In contrast, the skills of lava avoidance, goal-reaching, and moving obstacle avoidance are not required for this type of environment.
In this environment, we employ Proximal Policy Optimization (PPO) with $25600$ steps per policy update. The batch size used for each update is set to $64$, and an entropy coefficient of $0.01$ is applied. The image-based observation is flattened into a $1$D array, and a shared MLP policy and value network of $[256, 128, 64, 32]$ units is used. All the remaining parameters of PPO adopt the default setting of Stable Baselines 3~\cite{schulman2017proximal,stable-baselines3}. Furthermore, all the hyperparameters remain consistent across all the curriculum strategies evaluated.

\textbf{Bipedal Walker Stump Tracks (\textsc{BW}~\cite{romac2021teachmyagent}).} The experiment with \textsc{BW} is based on the stump tracks environment with a classic bipedal walker embodiment, as can be found in TeachMyAgent benchmark~\cite{romac2021teachmyagent}. The experimental setting is considered a mostly trivial task space, and the curriculum technique (teacher component) has no prior knowledge. Namely, in this setting, no reward mastery range, no prior knowledge concerning the task space, i.e., regions containing trivial tasks, and no subspace of test tasks are given. The learned policy that controls the walker agent with motor torque is trained with the Soft Actor Critic (SAC) algorithm for $20$ million steps.

\subsection{Curriculum Strategies Evaluated}\label{sec-app:exp-met-currs}

\textbf{Variants of our curriculum strategy.} Below, we report the hyperparameters and implementation details of the variants of our curriculum strategies used in the experiments:
\begin{enumerate}
\item For both \AlgoOurs{} and \AlgoOursUn{}:
\begin{enumerate}
\item $\beta$ parameter controls the softmax selection's stochasticity: we set $\beta = 90$ for the \textsc{SGR} and the \textsc{BW} environment, $\beta = 110$ for the \textsc{MiniG} environment, and $\beta = 130$ for \textsc{PM-s:1T} and \textsc{PM-s:2G} environments.
\item  $V_\textnormal{max}$ normalization parameter: we set $V_\textnormal{max} = 1$ for all environments. Since \textsc{BW} is a dense reward environment, the reward is scaled with the upper and lower bound rewards as provided by \cite{romac2021teachmyagent}.
\item $N_\text{pos}$ parameter that controls the frequency at which $V^t(\cdot)$ is updated: we set $N_\text{pos} = 5120$ for \envSGR{}, \textsc{PM-s:1T} and \textsc{PM-s:2G} environments and $N_\text{pos} = 25600$ for the \textsc{MiniG} environment, which is equivalent to the PPO update frequency. For \textsc{BW}, the frequency of updates is after each episode.
\end{enumerate}
\end{enumerate}

\textbf{State-of-the-art baselines.} Below, we report the hyperparameters and implementation details of the state-of-the-art curriculum strategies used in the experiments:
\begin{enumerate}
\item For \textsc{SPDL}~\cite{klink2020selfdeep}:
\begin{enumerate}
\item $V_\text{LB}$ performance threshold: we set $V_\text{LB} = 0.5$ for \textsc{SGR}, \textsc{PM-s:1T}, \textsc{PM-s:2G}, and \textsc{MiniG}.
\item $N_\text{pos}$ parameter that controls the frequency of performing the optimization step to update the distribution for selecting tasks: we set $N_\text{pos} = 5120$ for \textsc{SGR}, \textsc{PM-s:1T}, and \textsc{PM-s:2G}, and $N_\text{pos} = 25600$ for the \textsc{MiniG}.
\item For \textsc{BW}, we perform the experiments provided in~\cite{romac2021teachmyagent} for the Self-Paced teacher, which is equivalent to \textsc{SPDL} technique.
\end{enumerate}
\item For \textsc{CURROT}~\cite{klink2022curriculum}:
\begin{enumerate}
\item $V_\text{LB}$ performance threshold: we set $V_\text{LB} = 0.4$ for \envSGR{} and \textsc{MiniG}, $V_\text{LB} = 0.6$ for \textsc{PM-s:1T} and \textsc{PM-s:2G}, and $V_\text{LB} = 180$ for \textsc{BW}.
\item $\epsilon$ distance threshold between subsequent distributions: we set $\epsilon = 1.5$ for \envSGR{} and \textsc{MiniG}, $\epsilon = 0.05$ for \textsc{PM-s:1T} and \textsc{PM-s:2G}, and $\epsilon = 0.5$ for \textsc{BW}.
\item We choose the best-performing pair $(V_\text{LB}, \epsilon)$ for each environment from the set $\{0.4, 0.5, 0.6\} \times \{0.05, 0.5, 1.0, 1.5, 2.0\}$. For \textsc{BW}, we use the hyperparameters provided in \cite{klink2022curriculum}. 
\item $N_\text{pos}$ parameter that controls the frequency of performing the optimization step to update the distribution for selecting tasks: we set $N_\text{pos} = 5120$ for \textsc{SGR}, \textsc{PM-s:1T}, and \textsc{PM-s:2G}, and $N_\text{pos} = 25600$ for the \textsc{MiniG}.
\item The implementation in this paper incorporates all the original components of the strategy, including the update of the success buffer, the computation of the updated context distribution, and the utilization of a Gaussian mixture model to search for contexts that meet the performance threshold.
\item At the beginning of the training process, the initial contexts are uniformly sampled from the context space $\mathcal{C}$, following the same approach utilized in all other techniques.
\end{enumerate}
\item For \textsc{PLR}~\cite{jiang2021prioritized}:
\begin{enumerate}
\item $\rho$ staleness coefficient: we set $\rho = 0.5$ for the \textsc{SGR}, \textsc{PM-s:1T} and \textsc{PM-s:2G}, and \textsc{MiniG}.
\item $\beta_\text{PLR}$ temperature parameter for score prioritization:  we set $\beta_\text{PLR} = 0.1$ for all the environments.
\item $N_\text{pos}$ parameter that controls the frequency at which $V^t(\cdot)$ is updated: we set $N_\text{pos} = 5120$ for \textsc{SGR}, \textsc{PM-s:1T}, and \textsc{PM-s:2G} environments and $N_\text{pos} = 25600$ for the \textsc{MiniG} environment.
\item The technique has been adapted to operate with a pool of tasks. By employing a binomial decision parameter $d$, a new, unseen task is randomly selected from the task pool and added to the set of previously seen tasks. The seen tasks are prioritized and chosen based on their learning potential upon revisiting, aligning with the approach utilized in the original strategy.  As more unseen tasks are sampled from the pool, the binomial decision parameter $d$ is gradually annealed until all tasks are seen. Once this occurs, only the seen tasks are sampled from the replay distribution, taking into account their learning potential.
\end{enumerate}
\item For \textsc{Gradient}~\cite{huang2022curriculum}:
\begin{enumerate}
    \item Number of stages $N_\textnormal{stage}$: we set $5$ stages for \envSGR{}, $7$ stages for \textsc{MiniG} and $10$ stages for \textsc{PM-s:1T}, \textsc{PM-s:2G}, and \textsc{BW}. We selected the number of stages based on the original paper experiments and a value search in the set  $\{3,5,7,10\}$.
    \item Maximum number of steps per stage: we select $100000$ steps as the maximum number of training steps before switching to the next stage for \textsc{SGR}, \textsc{PM-s:1T}, and \textsc{PM-s:2G} environments. For \textsc{MiniG} we select $450000$ steps as the maximum number of steps per stage.
     \item $\Delta\alpha_{\textsc{gradient}}$ per stage: we set $\Delta\alpha = 0.2$ for \envSGR{}, and $\Delta\alpha=0.1$ for \textsc{BW}. For \textsc{PM-s:1T}, \textsc{PM-s:2G}, and \textsc{MiniG}, we choose the next $\alpha$ based on $\alpha(i)=\frac{1}{N_\textnormal{stage} - i}$. These selections are based on the experimental section and the provided implementation from \cite{huang2022curriculum}.
    \item Reward threshold per stage is set to $0.8$ for \envSGR{}, \textsc{PM-s:1T}, \textsc{PM-s:2G}, and \textsc{BW}. For \textsc{MiniG} the reward threshold is set to $0.4$. If the policy achieves this threshold, it switches to the next stage before reaching the maximum number of steps.
    
\end{enumerate}
\end{enumerate}

Our curriculum strategy only requires forward-pass operation on the critic-model to obtain value estimates for a subset of tasks $c$ and $c_{\text{targ}}$ in $\mathcal{C}$, followed by an $\arg\max$ operation over this subset. We note that the computational overhead of our curriculum strategy is minimal compared to the baselines. In particular, \textsc{SPDL} and \textsc{CURROT} require the same forward-pass operations and perform an additional optimization step to obtain the next task distribution. \textsc{CURROT} relies on solving an Optimal Transport problem requiring a high computational cost. Even when reducing the Optimal Transport problem to a linear assignment problem, as done in practice, the complexity is $O(n^3)$. As for \textsc{PLR}, it has an additional computational overhead for scoring the sampled tasks.